\documentclass[12pt]{article}
\usepackage{amsmath}
\usepackage{amssymb}
\usepackage{amsthm}
\usepackage{graphicx}
\usepackage{enumerate}
\usepackage{natbib}
\usepackage{tabularray}
\usepackage{url} 

\usepackage{amsfonts}
\usepackage{amsthm}
\usepackage{multirow}   
\usepackage{setspace}
\usepackage{amssymb}
\usepackage{amsmath}
\usepackage{graphicx,color,epstopdf}
\usepackage{hyperref}
\usepackage{xcolor}
\usepackage{booktabs}
\usepackage{subfigure}
\usepackage{array}

\usepackage{dsfont }  
\usepackage{algorithm}  
\usepackage{algorithmic}
\usepackage{enumerate}     
\usepackage{multirow}
\usepackage{setspace}

\usepackage{float}

\usepackage{graphicx}

\usepackage{threeparttable}

\newtheorem{theorem}{Theorem}[section]

\newtheorem{coro}{Corollary}[section]
\newtheorem{lemma}{Lemma}[section]
\newtheorem{remark}{Remark}[section]

\newtheorem{assumption}{Assumption}[section]

\numberwithin{equation}{section}

\begin{document}
	\title{Distributed quasi-Newton robust estimation under differential privacy}
 \author{Chuhan Wang$^{1}$,  Lixing Zhu$^{1}$ and Xuehu Zhu$^{2}$\thanks{
	 		Lixing Zhu is the corresponding author. Email: lzhu@bnu.edu.cn. The coauthors' names are in alphabetical order. The research described herewith was supported by a grant (NSFC12131006) from  the National Natural Scientific Foundation of China. ??? }\\
    \textit{$^1$ Department of Statistics, Beijing Normal University at Zhuhai, China}\\
$^2$ School of Mathematics, Xi'an Jiaotong University, Xi'An, China}   
    \date{}
	\maketitle
	\begin{abstract}
For distributed computing with Byzantine machines under Privacy Protection (PP) constraints, this paper develops a robust PP distributed quasi-Newton estimation, which only requires the node machines to transmit five vectors  to the central processor with high asymptotic relative efficiency.  Compared with the gradient descent strategy which requires more rounds of transmission and the Newton iteration strategy which requires the entire Hessian matrix to be transmitted, the novel quasi-Newton iteration has advantages in reducing privacy budgeting and transmission cost.  Moreover, our PP algorithm does not depend on the boundedness of gradients and second-order derivatives. When gradients and second-order derivatives follow  sub-exponential distributions, we offer a mechanism  that can ensure PP with  a sufficiently high probability. 
 Furthermore,  this novel estimator can achieve the optimal convergence rate and the asymptotic normality. The numerical studies on synthetic and real data sets evaluate the performance of the proposed algorithm.
	\end{abstract}
 
	\section{Introduction}
	In recent years, with the decreasing cost of data collection and storage, massive data has become a common concern in statistical analysis.  However, the data utilized for training models is often sensitive and possessed by private individuals or companies. 
Although data analysts can make the most accurate statistical inferences by directly obtaining raw data from data owners, this may put individuals at risk of leaking sensitive information. Accordingly, it is necessary to construct information transmission methods that fulfill the requirements of data analysis while ensuring stringent privacy protection for sensitive data. 

 In practical problems, massive personal data sets often come from multiple sources. For privacy protection purposes, data analysts cannot obtain all data from these sources, while only certain statistics for analysis.  This can be seen as a distributed computing framework: the node machines are responsible for storing data and transmitting privacy-preserving  statistics to the central processor, and the central processor is responsible for computing and providing necessary information feedback to the node machines.  We are motivated to develop a distributed computing method that protects individual privacy with high probability and makes statistical inferences as accurately as possible. 


 Besides, due to machine crashes, interference during data transmission or other reasons, some node machines may behave abnormally, or even exhibit Byzantine failures—arbitrary and potentially adversarial behavior. 
 To handle this situation, we propose the robust algorithm such that when a small number of machines encounter anomalies or  transmission problems, the final estimation results still have good statistical properties. This paper mainly focuses on developing robust  M-estimation methods under privacy protection in a distributed framework.
	
	\subsection{Problem formulation}

		A general M-estimation problem is formulated as follows:
	\begin{align}\label{Me1}
		\boldsymbol{\theta^*}=\underset{\boldsymbol{\theta} \in\boldsymbol{\Theta} }{\operatorname{argmin}} \mathbb{E}_{\boldsymbol{X}}\{f(\boldsymbol{X}, \boldsymbol{\theta})\}=\underset{\boldsymbol{\theta} \in\boldsymbol{\Theta} }{\operatorname{argmin}} \int_{\mathcal{X}}f(\boldsymbol{X}, \boldsymbol{\theta})dF(\boldsymbol{x}),
	\end{align}
	where $\boldsymbol{X}=(X_1,X_2,\cdots,X_q)^{\top}$ is a $p$-dimensional random vector in the sample space $\mathcal{X}\subset\mathbb{R}^p$, and $\boldsymbol{\theta}=(\theta_1,\theta_2,\cdots,\theta_p)^{\top}$ is a $p$-dimensional vector in the parameter space $\boldsymbol{\Theta}\in \mathbb{R}^p$. Moreover, $f$ is a convex loss function, and $F(\boldsymbol{x})$ represents the cumulative distribution function of  $\boldsymbol{X}$. 
	
	In a distributed setting, we assume there are totally $N=n(m+1)$ i.i.d. samples $\{\boldsymbol{X}_1,\boldsymbol{X}_2,\cdots,\boldsymbol{X}_N\}$ evenly stored on $m+1$ different machines $\{\mathcal{I}_0,\mathcal{I}_1,\mathcal{I}_2,\cdots\mathcal{I}_m\}$, including one central processor $\mathcal{I}_0$ and $m$ node machines $\{\mathcal{I}_1,\mathcal{I}_2,\cdots\mathcal{I}_m\}$, and each machine has $n$ samples. Some distributed M-estimation methods have been in the literature. For example, \cite{Jordan2019Communication}  
	transforms parameter estimation into a gradient descent problem, fully utilizing the gradients of all data to achieve the optimal convergence rate in a statistical sense. 
	Besides, \cite{huo2019distributed}  reduces the rounds of communication by passing the Hessian matrix,  \cite{fan2023communication} considers a distributed M-estimation method with penalty terms, and \cite{wang2023robust} studies the M-estimation problem when data on different machines has heterogeneity. 
	
	For robust estimation in a distributed framework, we usually refer to machines that provide abnormal data as Byzantine machines, and the resulting errors are called Byzantine failures \citep{1982The}. Concretely speaking, Byzantine machines may behave completely arbitrary and send unreliable information to the central processor.  Here we assume the proportion of Byzantine machines to be $\alpha_n$, and use $\mathcal{B}\subset\{1,\cdots,m\}$ to represent the index set of the Byzantine machines.

     The key to solve  Byzantine failures is adopting robust estimation strategies.  The methods in the literature include median estimation, 
	 trimmed mean estimation \citep{2018Byzantine, 2019Defending}, geometric median estimation \citep{chen2017distributed}  and iterative filtering estimation \citep{su2019securing}, 
 and among these methods,  median and trimmed mean estimation are the most commonly used. However, these two methods have low asymptotic relative efficiency. For example, compared with the mean estimation of normal samples, the asymptotic relative efficiency of the median estimation is only $0.637$, and the trimmed mean  estimation is $1-\beta$, where $\beta\geq 2\alpha_n$ is the proportion of truncated samples. Therefore, we try to construct robust statistics with higher asymptotic relative efficiency.
	 
	
	In terms of privacy protection issues,  the currently recognized privacy protection mechanisms mainly include the $(\varepsilon,\delta)$-differential privacy protection \citep{wang2024differentially, cai2024optima} and the f-differential privacy protection \citep{su2022dp, awan2023canonical}. Compared to the f-differential privacy protection, the differential privacy mechanism is simpler and easier to implement, and it is widely used in practical problems. There are three main mechanisms for differential privacy protection by adding noise to statistics: the Laplace mechanism, the Gaussian mechanism, and the exponential mechanism. Laplace and Gaussian mechanisms are simpler then the exponential mechanism, and Gaussian mechanism has better performances under the f-differential privacy protection framework.

	\subsection{Contributions}
	To address Byzantine failures and privacy protection issues for distributed M-estimation, while minimizing communication costs between machines and ensuring the effectiveness of the algorithm, we  adopt the ideas of quasi-Newton iteration and composite quantile estimation (\cite{Goldfarb} and \cite{2008Composite}), and achieve differential privacy protection by Gaussian mechanism to the transmitted statistics. The proposed method has the following features.
	
	(1) For general Newton iteration (see, e.g., \cite{huo2019distributed}, when transmitting the Hessian matrix, it needs to add the noise to every element of the $p\times p$ Hessian matrix for privacy protection. When the dimension $p$ is large, the privacy budget required by the Hessian matrix is much greater than that required by the gradient.  In other words, to achieve the same level of privacy protection as transmitting a $p$-dimensional vector, a greater amount of noise must be added to the Hessian matrix. This causes the accuracy loss of the final estimator. However,  we will see that the quasi-Newton method only requires the transmission of a $p$-dimensional vector, with the noise added to its $p$ coordinates. As a result, the privacy budget of each iteration is close to the gradient descent, and the quasi-Newton algorithm requires less vector transmission. The proposed robust quasi-Newton algorithm can reach the optimal convergence rate after transmitting five vectors and iterating two rounds. It is worth noting that the privacy budget will also increase linearly with the times of vector transmission, therefore, under the same estimation accuracy, the quasi-Newton method usually has less privacy budget than the Newton iteration. 	Moreover, in the second iteration, the quasi-Newton method avoids the difficulty of recomputing the inverse matrix, which is particularly beneficial when $p$ is large. 
	
	(2) In terms of robust estimation strategy, inspired from  the idea of composite quantile estimations widely used to reduce variance in \citep{2008Composite, lin2019composite, hou2023sparse}, 
	we suggest a  Distributed Composite Quantile (DCQ) estimator to handle the statistics used for iteration, including the gradients, and the Hessian matrices.
	As long as the estimators computed by node machines converge to the normal distribution, 
	we can apply the DCQ method to these local estimators for the final robust estimation. 
 The main advantage of the DCQ estimations is that they have much higher asymptotic relative efficiency than the median estimation. Specifically, for normal samples, the asymptotic relative efficiency of the median estimator relative to the mean estimator is $0.653$, whereas the DCQ estimator can reach $0.955$. 

	(3) We use the Gaussian mechanism for the differential privacy protection instead of the Laplace mechanism. 
 The Gaussian mechanism can maintain these limiting normal distributions with a slightly larger variance, satisfying the requirements of the DCQ estimations. In contrast, Laplace noise disrupts the normality of the original local estimators, 
	which may slow down the theoretical convergence rates of the estimators. Besides, the Gaussian mechanism can also achieve optimal $f$-differential privacy protection, see \cite{su2022dp} for details.

(4) In previous privacy-preserving strategies for parameter estimation, boundedness assumptions are often made to ensure bounded sensitivity. However, when dealing with statistics such as the gradients being generally unbounded, such assumptions significantly increase the limitations of the methods. The typical Gaussian distribution is also not applicable under boundedness assumptions. A more representative approach involves controlling the tail probability of statistics using sub-Gaussian or sub-exponential distribution assumptions. This allows the definition of sensitivity such that differential privacy protection can be achieved with a sufficiently high probability. \cite{cai2024optima} provides an illustrative example in the context of principal component analysis. Our research applies this approach primarily to M-estimators, gradient estimators, and other useful estimators in distributed iterative procedures, thereby relaxing the boundedness constraints for statistics in distributed optimization algorithms with privacy protection.

	
	 \subsection{Organization}

 This paper is organized as follows. Section 2 introduces the notations and the basic principle of the Gaussian privacy protection mechanism. Section 3 presents the Distributed Composite Quantile (DCQ) estimators. Section 4 develops a differentially private  robust  distributed quasi-Newton estimator and outlines its limiting properties. Section 5 discusses the simulation results and real data analysis. Section 6 includes some further discussions. Section 7 details the regularity assumptions. All proofs are provided in Supplementary Materials.
	
 \section{Preliminaries}
  \subsection{Notations}
(1) For any vector $\boldsymbol{v}=(v_1,v_2,\ldots,v_p)^{\top}$, define  $\|\boldsymbol{v}\|=(\sum_{l=1}^{p}v_l^2)^{1/2}$ and $\boldsymbol{v}^{\otimes 2}=\boldsymbol{v}\boldsymbol{v}^{\top}$. For two vectors $\boldsymbol{u}\in\mathbb{R}^p$ and $\boldsymbol{v}\in \mathbb{R}^p$, let $\langle\boldsymbol{u},\boldsymbol{v}\rangle=\sum_{l=1}^{p}u_lv_l$.

(2) For a matrix $\mathbf{M}$, denote $\|\mathbf{M}\|=\sup_{\|\boldsymbol{a}\|=1}{\|\boldsymbol{a}^{\top}\mathbf{M}\|}$ as the spectral norm, which is the largest eigenvalue of $\mathbf{M}$ when $\mathbf{M}$ is a square matrix. Denote $\operatorname{Tr}(\mathbf{M})$ as the trace of $\mathbf{M}$, and let $\mathbf{M}_{l\cdot}$ represent the row vector of the $l$-th row of $\mathbf{M}$. 
Write the largest and smallest eigenvalues of $\mathbf{M}$ as $\lambda_{\max}(\mathbf{M})$ and $\lambda_{\min}(\mathbf{M})$, respectively. Let $\operatorname{diag}(\mathbf{M})$ be the vector of entries on the diagonal of $\mathbf{M}$.

(3) Define $B(\boldsymbol{v},r)=\{\boldsymbol{w}\in\mathbb{R}^p:|\boldsymbol{w}-\boldsymbol{v}|\leq r\}$. Let $\mathbb{I}(\cdot)$ be the indicator function.

(4) Write $F^{-1}(y)$ as the smallest $x\in\mathbb{R}$ satisfying $F(x)\geq y$. Let $\mathbf{N}(0,1)$ denote the standard normal distribution with the distribution function $\Psi(x)=\mathbb{P}(\mathbf{N}(0,1)\leq x)$ and the density function $\psi(x)=\frac{1}{\sqrt{2\pi}}\exp(-\frac{x^2}{2})$.

(5) Use $\nabla$ and $\nabla^2$ to represent the gradient and Hessian operators. For $f(\boldsymbol{\theta})=f(\theta_1,\theta_2,\ldots,\theta_p)$, $\nabla f_l$ denotes the partial derivative with respect to $\theta_l$, and $\nabla^2 f_{l_1l_2}$ represents the second-order derivatives of $\theta_{l_1}$ and $\theta_{l_2}$.  Moreover, $\nabla f$ and $\nabla^2 f$ refer to  the vector with the $l$-th entry $\nabla f_l$ and the matrix with the $(l_1,l_2)$-entry $\nabla^2 f_{l_1l_2}$, respectively..

(6) For any positive integer $N$, denote $[N]$ as the index set $\{1,2,\ldots,N\}$ and $[N]_0$ as the index set $\{0,1,2,\ldots,N\}$. For two sets $\mathcal{A}_1$ and $\mathcal{A}_2$, denote $\mathcal{A}_1\backslash\mathcal{A}_2$ as the set of elements belonging to $\mathcal{A}_1$ but not to $\mathcal{A}_2$.

(7) For $X_1,X_2,\ldots,X_m$, let $\operatorname{med}\{X_j,j\in[m]\}$ be the median of $\{X_1,X_2,\ldots,X_m\}$. For $m$ vectors $\boldsymbol{a}_j=(a_{j,1},a_{j,2},\ldots,a_{j,p})^{\top}$, $j\in[m]$, define $\operatorname{med}\{a_j,j\in[m]\}=(\operatorname{med}\{a_{j,1},j\in[m]\},\ldots,\operatorname{med}\{a_{j,p},j \in [m]\})^{\top}$, where $\operatorname{med}\{a_{j,l},j\in[m]\}$ is the median of $\{a_{j,l},j\in[m]\}$, $l\in[p]$.

(8) For M-estimation, write the $m+1$ machines as $\mathcal{I}_0$, $\mathcal{I}_1$, $\mathcal{I}_2$, \ldots, $\mathcal{I}_m$, where $\mathcal{I}_0$ acts as the central processor, and the others are node machines. For any random variable $\boldsymbol{X}$, let $\mathbb{E}_{\boldsymbol{X}}(\cdot)$ represent the expectation over $\boldsymbol{X}$. For $j\in[m]_0$, define $F_j(\boldsymbol{\theta})=\frac{1}{n}\sum_{i \in \mathcal{I}_j} f\left(\boldsymbol{X}_i, \boldsymbol{\theta}\right)$. Let $\boldsymbol{\hat{\theta}}_j=(\hat{\theta}_{j 1}, \ldots, \hat{\theta}_{j p})=\operatorname{argmin}\limits_{\boldsymbol{\theta}\in \boldsymbol{\Theta}} F_j(\boldsymbol{\theta})$. Define the global loss function $F(\boldsymbol{\theta})=\frac{1}{m+1} \sum_{j=0}^{m} F_j(\boldsymbol{\theta})=\frac{1}{(m+1)n}\sum_{i=1}^{(m+1)n}f(\boldsymbol{X}_i,\boldsymbol{\theta})$ and $F_{\mu}(\boldsymbol{\theta})=\mathbb{E}\{f(\boldsymbol{X},\boldsymbol{\theta})\}$.

 \subsection{A brief review of privacy framework}

In a distributed framework, allowing the central processor to access all original data on each node machine is unrealistic and poses privacy risks. Differential privacy mechanisms are typically used to safeguard privacy, ensuring that each node machine protects its local data and communicates only privatized information to the central processor for further analysis.

	Differential Privacy (DP) was first conceptually proposed by \cite{dwork2006calibrating}. If two data sets $\mathbf{X}$ and $\mathbf{X}'$ differ by only one datum, 
 we say $\mathbf{X}$ and $\mathbf{X}'$ are a pair of adjacent data set.	A randomized algorithm $\mathcal{M}:\mathcal{X}^n\to\mathbb{R}^p$ is called ($\varepsilon,\delta$)-differentially private if for every pair adjacent data sets $\mathbf{X},\mathbf{X}{'}\in\mathcal{X}^n$ that differ by one individual datum and every $S\in\mathbb{R}^p$,
	\begin{align*}
		\mathbb{P}(\mathcal{M}(\mathbf{X})\in S)\leq e^{\varepsilon}\cdot \mathbb{P}(\mathcal{M}(\mathbf{X}{'})\in S)+\delta,
	\end{align*}
	where the probability $\mathbb{P}$ is  only induced by the randomness of $\mathcal{M}$. 
	
	To provide a randomized algorithm that satisfies DP, we need to define $\ell_2$ sensitivity. The $\ell_2$-sensitivity of $\mathcal{M}:\mathbf{X}^n\to\mathbb{R}^p$ is defined as $\Delta=\sup_{\mathbf{X},\mathbf{X}'}\|\mathcal{M}(\mathbf{X})-\mathcal{M}(\mathbf{X}')\|$, where $\mathbf{X}$ and $\mathbf{X}'$ are a pair of adjacent data sets.
	Then the following Gaussian mechanism can achieve $(\varepsilon,\delta)$-DP.
	
	\begin{lemma}\label{g1}(\cite{dwork2014algorithmic})
Given a function $\mathcal{M}:\mathcal{X}^n\to\mathbb{R}^p$ with the $\ell_2$-sensitivity $\Delta$ and a dataset $\mathbf{X}\subset\mathcal{X}^n$,  assume $\sigma\geq\frac{\sqrt{2\log(1.25/\delta)}\Delta}{\varepsilon}$. The following Gaussian mechanism yields $(\varepsilon,\delta)$-DP:
		\begin{align*}
			\mathcal{G}(\mathbf{X},\sigma):=\mathcal{M}(\mathbf{X})+\boldsymbol{b},\quad \boldsymbol{b}\sim\mathbf{N}(\boldsymbol{0},\sigma^2\mathbf{I}_p),
		\end{align*}
	where $\mathbf{I}_p$ is a $p\times p$ identity matrix.
	\end{lemma}

The Gaussian mechanism above requires a finite sensitivity value. However, in many cases, the sensitivity under strict definition is unbounded (e.g. $\mathcal{M}$ is the average of i.i.d. standard normal distribution random vectors) and then the Gaussian mechanism fails to hold. To solve this problem, most existing studies designed some bounded conditions, for example, \cite{zhou2020differentially} and \cite{wang2024differentially} assumed that the $L_2$ norm of the first derivative of the loss functions in their algorithms are bounded. However, if we assume sub-exponential (sub-Gaussian) distributions   for i.i.d. random variables, then for their average, by adjusting the noise variance, the Gaussian mechanism can achieve $(\varepsilon,\delta)$ privacy protection with a sufficiently high probability. We will discuss this issue in detail in Section 4.

\section{Distributed composite quantile estimation}
 In a distributed framework, for given $n$, let $\{Y_{j}^{(n)},j\in[m]\}$ be $m$ i.i.d. observations that are usually  summary statistics of a random variable $Y^{(n)}$. 
 When $Y^{(n)}$ converges to a symmetric random  variable $Y$ as $n\to\infty$, we can use $\hat{Y}_{med}^{(n)}=\operatorname{med}\{Y_j^{(n)},j\in[m]\}$ to estimate $\mathbb{E}(Y^{(n)})$, which is robust but has low asymptotic relative efficiency. Let $g(\cdot)$ and $G(\cdot)$ denote the density function and the distribution function of $Y-\mathbb{E}(Y)$, respectively.  To reduce the variance of $\hat{Y}_{med}^{(n)}$, we modify $\hat{Y}_{med}^{(n)}$ by using the idea of composite quantile estimation (\cite{2023byzantine}): for a given integer $K$, 
    \begin{align}\label{eq21}
		\hat{Y}_{cq}^{(n)} =&\hat{Y}_{med}^{(n)}-\frac{\hat{\sigma}_{Y^{(n)}}\sum_{k=1}^{K}\sum_{j=1}^{m}\left[\mathbb{I}(Y_j^{(n)}\leq \hat{Y}_{med}^{(n)}+\hat{\sigma}_{Y^{(n)}}\Delta_k)-\kappa_k\right]}{m\sum_{k=1}^{K}g(\Delta_k)},
    \end{align}
    where $\kappa_k=\frac{k}{K+1}$,  $\Delta_k=G^{-1}(\kappa_k)$ and $\hat{\sigma}_{Y^{(n)}}^2$ is the  variance estimator of $Y^{(n)}$. Compared to $\hat{Y}_{med}^{(n)}$, $\hat{Y}_{cq}^{(n)}$ utilizes more information about  the empirical distribution of $Y^{(n)}$.  We will define the estimator with the form in (\ref{eq21}) as the Distributed Composite Quantile (DCQ) estimator. 
  The following Theorem \ref{th21} guarantees the convergence rate of $\hat{Y}_{cq}^{(n)}$. 
\begin{theorem}\label{th21}(Convergence rate of the DCQ estimator) 
	Let $\alpha_n$ be the proportion of Byzantine machines and $\hat{Y}_{cq}^{(n)}$ be defined in (\ref{eq21}).  Suppose that $\{Y_j^{(n)},j\in[m]\}$ are  i.i.d random variables with 
	\begin{align*}
		\left|\mathbb{P}\left((Y_j^{(n)}-\mu_Y)/\sigma_{Y^{(n)}}<x\right)-G(x)\right|=O(\rho_n),
	\end{align*}
 where $\mu_Y$ is a constant and $\sigma^2_{Y^{(n)}}=\operatorname{Var}(Y^{(n)})$. 
	Moreover, assume $G(\cdot)$ satisfy Assumption \ref{gf} in Section~7 and $\hat{\sigma}_{Y^{(n)}}-\sigma_{Y^{(n)}}=O_p(\eta_n)$.
	Then
	\begin{align*}
		\hat{Y}_{cq}^{(n)}-\mu_Y=O_p\left(\frac{1}{\sqrt{m}}+\frac{\log n}{m}+\sqrt{\frac{(\alpha_{n}+\rho_n+\eta_n) \log n}{m}}+\alpha_n+\rho_n+\eta_n\right).
	\end{align*}
	When $\log n=o(\sqrt{m})$ and $\alpha_n+\rho_n+\eta_n=o(1/\sqrt{m})$, we have
	\begin{align}\label{li1}
		{\sqrt{m}(\hat{Y}_{cq}^{(n)}}-\mu_Y)/{\sigma_{cq}}\stackrel{d}{\to}\mathbf{N}(0,1),
	\end{align}
 where $\sigma_{cq}^2=D_K\sigma_{Y^{(n)}}^2$ with $D_K=\sum_{k_1=1}^{K}\sum_{k_2=1}^{K}[\min\{\kappa_{k_1},\kappa_{k_2}\}/{\{\sum_{k=1}^{K}\psi(\Delta_{k})\}^2}]$.
	
\end{theorem}

\begin{remark}
Equation (\ref{li1}) in Theorem \ref{th21} shows that for $m$ local estimators computed by the node machines  converging to a certain distribution, if $\alpha_n$, $\rho_n$ and $\eta_n$ satisfy some rate conditions, then the DCQ estimator of these local estimators  can  converge to normal distribution with the rate $\sqrt{m}$. For example, let $\{X_{j,i},i\in[n],j\in[m]_0\}$ be i.i.d. random variables with mean $\mu_X$, $Y_j^{(n)}=\frac{1}{\sqrt{n}}\sum_{i=1}^{n}(X_{j,i}-\frac{1}{n}\sum_{i=1}^{n}X_{j,i})$, $\hat{\sigma}_{Y^{(n)}}=\frac{1}{n}\sum_{i=1}^{n}(X_{0,i}-\frac{1}{n}\sum_{i=1}^{n}X_{0,i})^2$,  $\alpha_n=\frac{1}{\sqrt{m}}$ and $\log n=o(m)$, then $\rho_n=\eta_n=\frac{1}{\sqrt{n}}$ and $\hat{Y}_{cq}^{(n)}-\mu_X=O_p(\frac{1}{\sqrt{m}})$. In this example, the DCQ estimator degenerates into the VRMOM estimator proposed by \cite{tu2021variance}. Furthermore,  the DCQ estimation can be applied to more statistics,  including the M-estimation.
\end{remark}
	\section{ Differentially Private Robust Distributed Quasi-Newton Estimator}
In Subsection \ref{gn0}, we present the computation of the differentially private robust distributed quasi-Newton estimator step by step.  To maintain content coherence, we defer the discussion of the differential privacy guarantee to Subsection \ref{gn1}. Furthermore, Subsection \ref{UCP} examines the scenario in which the central processor may act as a Byzantine machine.

 \subsection{ Quasi-Newton algorithm}\label{gn0}
The quasi-Newton method does not compute the Hessian matrix directly. Instead, it approximates the Hessian or its inverse, while retaining certain properties of the Hessian that provide essential information for determining the gradient descent direction. Depending on how the Hessian matrix is approximated, common quasi-Newton methods are categorized into three types: Symmetric Rank-One update (SR1) \cite{Davidon}, Broyden-Fletcher-Goldfarb-Shanno update (BFGS) \cite{Goldfarb}, and Davidon-Fletcher-Powell update (DFP) \cite{DFP}. Compared to SR1, both BFGS and DFP ensure the positive definiteness of the Hessian matrix estimation. There is a dual relationship between DFP and BFGS; however, BFGS often performs better than DFP in practice. Several studies have applied BFGS to distributed computing without considering robustness and privacy protection, such as \cite{chen2014bfgs}, \cite{eisen2017decentralized}, and \cite{wu2023quasi}. Thus, we recommend BFGS for quasi-Newton iteration. Notably, the computational cost of determining a general Hessian inverse matrix is $O(p^3)$. However, by approximating the Hessian inverse during the iteration without directly inverting the matrix, the quasi-Newton method reduces the computational complexity to $O(p^2)$ (\cite{wu2023quasi}).

We then briefly review the BFGS algorithm in a distributed framework.  Let $\boldsymbol{\hat{\theta}}_{(t)}$  be the parameter estimator after $t$  iterations.  $\nabla F(\boldsymbol{\hat{\theta}}_{(t)})$ and $\nabla^2 F(\boldsymbol{\hat{\theta}}_{(t)})$ are the corresponding derivative and Hessian matrix respectively. According to Taylor's expansion, 
    \begin{align*}
    	\nabla F(\boldsymbol{\hat{\theta}}_{(t+1)})-\nabla F(\boldsymbol{\hat{\theta}}_{(t)})\approx\nabla^2 F(\boldsymbol{\hat{\theta}}_{(t+1)})(\boldsymbol{\hat{\theta}}_{(t+1)}-\boldsymbol{\hat{\theta}}_{(t)}).
    \end{align*}
For the $j$-th machine, assume we already have $\mathbf{H}_j^{(t)}\approx\{\nabla^2 F(\boldsymbol{\hat{\theta}}_{(t)})\}^{-1}$. Then
    \begin{align}\label{app1}
    	\mathbf{H}_j^{(t+1)}\{\nabla F(\boldsymbol{\hat{\theta}}_{(t+1)})-\nabla F(\boldsymbol{\hat{\theta}}_{(t)})\}\approx \boldsymbol{\hat{\theta}}_{(t+1)}-\boldsymbol{\hat{\theta}}_{(t)}.
    \end{align}
Given $\mathbf{H}_j^{(t)}$, $\mathbf{H}_j^{(t+1)}$ is updated to be    $$\mathbf{H}_j^{(t+1)}=\mathbf{H}_j^{(t)}+\alpha_j \boldsymbol{v}_j^{(t)}\{\boldsymbol{v}_j^{(t)}\}^{\top}+\beta_j \boldsymbol{u}_j^{(t)}\{\boldsymbol{u}_j^{(t)}\}^{\top}$$ for some  coefficient $\alpha_j,\beta_j \in \mathbb{R}$ and $\boldsymbol{v}^{(t)},\boldsymbol{u}^{(t)}\in \mathbb{R}^p$.
    Note that when
    $\boldsymbol{v}_0^{(t)}=\cdots=\boldsymbol{v}_m^{(t)}=\boldsymbol{\hat{\theta}}_{(t+1)}-\boldsymbol{\hat{\theta}}_{(t)}$, $\boldsymbol{u}_j^{(t)}=\mathbf{H}_j^{(t)}\{\nabla F(\boldsymbol{\hat{\theta}}_{(t+1)})-\nabla F(\boldsymbol{\hat{\theta}}_{(t)})\}$,
    $\alpha_0=\cdots=\alpha_m=[\{\boldsymbol{v}_0^{(t)}\}^{\top}\{\nabla F(\boldsymbol{\hat{\theta}}_{(t+1)})-\nabla F(\boldsymbol{\hat{\theta}}_{(t)})\}]^{-1}$ and $\beta_j=-[\{\boldsymbol{u}_j^{(t)}\}^{\top}\{\nabla F(\boldsymbol{\hat{\theta}}_{(t+1)})-\nabla F(\boldsymbol{\hat{\theta}}_{(t)})\}]^{-1}$, the left-hand term and the right-hand term in (\ref{app1}) are equal for all $j\in[m]_0$.

    Thus, through a simple organization, we update $\mathbf{H}_j^{(t+1)}$  as:
    $$
    \mathbf{H}_j^{(t+1)}=(\mathbf{V}^{(t)})^{\top} \mathbf{H}_j^{(t)} \mathbf{V}^{(t)}+\rho^{(t)}(\boldsymbol{\hat{\theta}}_{(t+1)}-\boldsymbol{\hat{\theta}}_{(t)})(\boldsymbol{\hat{\theta}}_{(t+1)}-\boldsymbol{\hat{\theta}}_{(t)})^{\top},
    $$
    where $\rho^{(t)}=[(\boldsymbol{\hat{\theta}}_{(t+1)}-\boldsymbol{\hat{\theta}}_{(t)})^{\top}\{\nabla F(\boldsymbol{\hat{\theta}}_{(t+1)})-\nabla F(\boldsymbol{\hat{\theta}}_{(t)})\}]^{-1}$ and $\mathbf{V}^{(t)}=\mathbf{I}_p-\rho^{(t)}\{\nabla F(\boldsymbol{\hat{\theta}}_{(t+1)})-\nabla F(\boldsymbol{\hat{\theta}}_{(t)})\}(\boldsymbol{\hat{\theta}}_{(t+1)}-\boldsymbol{\hat{\theta}}_{(t)})^{\top}$. It is worth noting that both $\rho^{(t)}$ and $\mathbf{V}^{(t)}$ are unrelated to $j$. Therefore, the main goal of distributed computing is to merge the information provided by $\mathbf{H}_j^{(t)}$, or more specifically, to merge the product of $(\mathbf{V}^{(t)})^{\top} \mathbf{H}_j^{(t)} \mathbf{V}^{(t)}$ and the gradient for $j\in[m]_0$, which are  $p$-dimensional vectors.
    
	    

     In the following subsections \ref{ini}-\ref{sec4.1.3}, we discuss how to gain the robust initial estimator  and update it by the robust quasi-Newton algorithm for distributed computing under privacy protection constraints. 
	\subsubsection{The construction of an initial estimator}\label{ini}
   
Based on the construction of the DCQ estimator, we consider how to obtain a robust parameter estimator that can achieve privacy protection without iteration. Before information communication, every machine can compute the local parameter estimators based on their own data, that is, $\boldsymbol{\hat{\theta}}_j=\operatorname{argmin}\limits_ {\boldsymbol{\theta} \in\boldsymbol{\Theta}}\frac{1}{n}\sum_{i\in\mathcal{I}_j}f(\boldsymbol{X}_i, \boldsymbol{\theta})$ for $j\in[m]_0$. To achieve ($\varepsilon,\delta$)-DP,  every machine needs to add a Gaussian noise to $\boldsymbol{\hat{\theta}}_j$, 
     	\begin{align}\label{thetajdp}
\boldsymbol{\hat{\theta}}_{j,DP}=\boldsymbol{\hat{\theta}}_j+\boldsymbol{b}_j^{(1)}=(\hat{\theta}_{j1}^{DP},\cdots,\hat{\theta}_{jp}^{DP})^{\top}	\ {\rm{with}}\  \boldsymbol{b}_j^{(1)}\stackrel{i.i.d.}{\sim}\mathbf{N}(\boldsymbol{0},s_1^2\mathbf{I}_p)
\end{align}
and then sends $\boldsymbol{\hat{\theta}}_{j,DP}$  to the central processor.

The following lemma guarantees that the estimator in (\ref{thetajdp})  satisfying $(\varepsilon,\delta)$-DP has a normal weak limit: 
\begin{lemma}\label{le301}
	Suppose that Assumptions \ref{a1}-\ref{a3} and \ref{a52}-\ref{a9} hold. 
 Let $\hat{\theta}_{jl}^{DP}$ be the $l$-th entry of $\boldsymbol{\hat{\theta}}_{j,DP}$ defined in (\ref{thetajdp}).  For $j\notin\mathcal{B}$, there exists a positive constant $C_a$ such that
	\begin{align}\label{03031}
		\underset{-\infty<u<\infty}{\sup}\left|\mathbb{P}\left(\frac{\sqrt{n}(\hat{\theta}_{jl}^{DP}-\theta_l^*)}{\{\sigma_l^2(\boldsymbol{\theta^*})+ns_1^2\}^{1/2}}\leq u\right)-\Psi(u)\right|\leq C_a\left(\frac{p\log n}{\sqrt{n}}\right).
	\end{align}
\end{lemma}
According to Theorem \ref{th21},  to compute the DCQ estimator of $\boldsymbol{\hat{\theta}}_{j,DP}$, we need to estimate the
	     variance of $\sqrt{n}{\hat{\theta}}_{jl}^{DP}$.
 Recall ${\boldsymbol{\hat{\theta}}}_{med,DP}=\operatorname{med}\{\boldsymbol{\hat{\theta}}_{j,DP},j\in[m]_0\}$. The variance of $\sqrt{n}{\hat{\theta}}_{jl}^{DP}$ can be estimated by $\hat{\sigma}_{bl}^2:=\hat{\sigma}_l^2({\boldsymbol{\hat{\theta}}}_{med,DP})+ns_1^2$, where 
\begin{align*}
	&\left(\hat{\sigma}_1^2({\boldsymbol{\hat{\theta}}}_{med,DP}),\cdots,\hat{\sigma}_p^2({\boldsymbol{\hat{\theta}}}_{med,DP})\right)\nonumber
	\\& =\operatorname{diag}\bigg(\{\nabla^2F_0(\boldsymbol{\boldsymbol{\hat{\theta}}}_{med,DP})\}^{-1}\frac{1}{n}\sum_{i\in\mathcal{I}_0}\{\nabla f(\boldsymbol{X}_i,\boldsymbol{\boldsymbol{\hat{\theta}}}_{med,DP})-\nabla F_0(\boldsymbol{\hat{\theta}}_{med,DP})\}^{\otimes 2} \\
 &\ \ \ \ \ \ \ \ \ \ \ \ \{\nabla^2F_0(\boldsymbol{\boldsymbol{\hat{\theta}}}_{med,DP})\}^{-1}\bigg).
\end{align*}
Lemma \ref{l10} shows that $\hat{\sigma}_l({\boldsymbol{\hat{\theta}}}_{med,DP})$ is a consistent estimator of $\sigma_{l}(\boldsymbol{\theta^*})$, and $ns_1^2$ is the increase for variance caused by Gaussian noise. Since this estimation only uses ${\boldsymbol{\hat{\theta}}}_{med,DP}$ and the sample on the central processor, this variance computation process does not add any additional information transmission.

  \begin{lemma}\label{l10}
	
Under Assumptions \ref{a1}-
	\ref{a9} and \ref{a11}-\ref{a12},  for any $l\in[p]$, 
	\begin{align*}	|\hat{\sigma}_l({\boldsymbol{\hat{\theta}}}_{med,DP})-\sigma_{l}(\boldsymbol{\theta^*})|=O_p(p/\sqrt{n}),
	\end{align*}
where $(\sigma_1^2(\boldsymbol{\theta^*}),\cdots,\sigma_p^2(\boldsymbol{\theta^*}))$ are the diagonal elements of $\boldsymbol{\Sigma}(\boldsymbol{\theta^*})=\{\nabla^2F_{\mu}(\boldsymbol{\theta^*})\}^{-1}\mathbb{E}\{\nabla f(\boldsymbol{X},\boldsymbol{\theta^*})^{\otimes 2}\} \\\{\nabla^2F_{\mu}(\boldsymbol{\theta^*})\}^{-1}$.
\end{lemma}
 Then we can use the DCQ estimator of $\{\hat{\theta}_{jl}^{DP},j\in[m]_0\}$ as the $l$-th entry of the initial estimator. Let $\boldsymbol{\hat{\theta}}_{cq,DP}=(\hat{\theta}_{cq,1}^{DP,K},\cdots,\hat{\theta}_{cq,p}^{DP,K})^{\top}$, where 
	    	\begin{align}\label{eq31}
	\hat{\theta}_{cq,l}^{DP,K}=&\hat{\theta}_{med,l}^{DP}-\frac{\hat{\sigma}_{bl}(\boldsymbol{\hat{\theta}}_{med,DP})\sum_{k=1}^{K}\sum_{j=1}^{m}\left[\mathbb{I}(\hat{\theta}_{jl}^{DP}\leq \hat{\theta}_{med,l}^{DP}+\hat{\sigma}_{bl}(\boldsymbol{\hat{\theta}}_{med,DP})\Psi^{-1}(\kappa_k)/\sqrt{n})-\kappa_k\right]}{m\sqrt{n}\sum_{k=1}^{K}\psi(\Psi^{-1}(\kappa_k))}
	    	\end{align}
      and  $\kappa_k=\frac{k}{K+1}$. The following theorem shows its convergence rate.
	    \begin{theorem}\label{th31}
	    	Suppose Assumptions \ref{a1}-\ref{a3}, \ref{a52}-\ref{a9} and \ref{a12} hold. 
	    	If $\alpha_n=O(1/\log n)$, ${\log^3 n}/{m}=o(1)$ and ${p^2\log^2 n}/{n}=o(1)$, then
	    \begin{align}\label{t31}
	    	\|\boldsymbol{\hat{\theta}}_{cq,DP}-\boldsymbol{\theta^*}\|=O_{p}\left(\frac{\alpha_n\sqrt{p}}{\sqrt{n}}+\frac{\sqrt{p}}{\sqrt{mn}}+\frac{p^{3/2}\log n}{n}\right).
	    \end{align}
	    \end{theorem}
	    When $p$ is relatively large compared to $n$ (e.g. $p^2\log^3 n=n$), the last term ${p^{3/2}\log n}/{n}$ in (\ref{t31}) may converge to zero slowly. We can improve the convergence rate through further iterations.

     

	    \subsubsection{One-stage estimator}
	    In this subsection, we discuss how to use gradients and Hessian matrices to update the initial estimator in the first iteration. After computing $\boldsymbol{\hat{\theta}}_{cq,DP}$, the central processor sends $\boldsymbol{\hat{\theta}}_{cq,DP}$ to the node machines, and then the node machines  compute $\nabla F_j(\boldsymbol{\hat{\theta}}_{cq,DP})$ by using their own samples. To achieve $(\varepsilon,\delta)$-DP, the node machines send 
     \begin{align}\label{eq4.5}
         \nabla F_j^{DP}(\boldsymbol{\hat{\theta}}_{cq,DP})=\nabla F_j(\boldsymbol{\hat{\theta}}_{cq,DP})+\boldsymbol{b}_j^{(2)} \ {\rm{with}} \ \boldsymbol{b}_j^{(2)}\sim\mathbf{N}(\boldsymbol{0},s_2^2\mathbf{I}_p)
     \end{align}
     to the central processor $\mathcal{I}_0$. Then the central processor uses these local gradients to compute their DCQ estimator $\widehat{\nabla F_{cq}^{DP}}(\boldsymbol{\hat{\theta}}_{cq,DP})$ and sends it back to the node machines. After that, the node machines are responsible for using their own data to compute the estimator of the inverse of the Hessian matrix, $\{\nabla^2 F_{j}(\boldsymbol{\hat{\theta}}_{cq,DP})\}^{-1}$,  and  to add Gaussian noise to the vector $\{\nabla^2 F_{j}(\boldsymbol{\hat{\theta}}_{cq,DP})\}^{-1}\widehat{\nabla F_{cq}^{DP}}(\boldsymbol{\hat{\theta}}_{cq,DP})$, that is 
	    \begin{align}\label{it1}
	    	\boldsymbol{h}_j^{(1)}=\{\nabla^2 F_{j}(\boldsymbol{\hat{\theta}}_{cq,DP})\}^{-1}\widehat{\nabla F_{cq}^{DP}}(\boldsymbol{\hat{\theta}}_{cq,DP})+\boldsymbol{b}_j^{(3)}\ {\rm{with}}\ \boldsymbol{b}_j^{(3)}\sim\mathbf{N}(\boldsymbol{0},s_{3,j}^2\mathbf{I}_p).
	    \end{align}     
The node machines send $\boldsymbol{h}_j^{(1)}$ to the central processor. Finally, the central processor computes the DCQ estimator of $\boldsymbol{h}_j^{(1)}$, which is written as $\boldsymbol{H}_1$,   the one-stage estimator can be defined as:
 \begin{align}\label{osdp}
\boldsymbol{\hat{\theta}}_{os,DP}=\boldsymbol{\hat{\theta}}_{cq,DP}-\boldsymbol{H}_1.
\end{align}
	 \begin{remark}
	 When $p$ is large, the computational complexity to compute $\{\nabla^2 F_{j}(\boldsymbol{\hat{\theta}}_{cq,DP})\}^{-1}$ is $O(p^3)$. To reduce computational burden,  we can also use the quasi-Newton method to estimate $\{\nabla^2 F_{j}(\boldsymbol{\hat{\theta}}_{cq,DP})\}^{-1}$, and the details can be found in Algorithm 1 in \cite{wu2023quasi}. Since this estimation is performed on a single machine and does not involve robust estimation and privacy protection, we will not discuss it further.	
	 \end{remark}

 Lemmas 4.5 and 5.1 in Supplementary Materials show that
$\nabla F_{jl}^{DP}(\boldsymbol{\hat{\theta}}_{cq,DP})$ and $\boldsymbol{h}_j^{(1)}$ have normal weak limits. To derive the DCQ estimator, we still need to estimate their variance. Similar to Subsection \ref{ini}, the variance of their entries can be  estimated by using the data on the central processor.  Write $\nabla F_j^{DP}(\boldsymbol{\theta})$ as $(\nabla F_{j1}^{DP}(\boldsymbol{\hat{\theta}}_{cq,DP}),\cdots,\nabla F_{jp}^{DP}(\boldsymbol{\hat{\theta}}_{cq,DP}))^{\top}$, the variance of $\sqrt{n}\nabla F_{jl}^{DP}(\boldsymbol{\hat{\theta}}_{cq,DP})$ can be estimated by $$\frac{1}{n}\sum_{i\in\mathcal{I}_0}\{\nabla f_l(\boldsymbol{X}_i,\boldsymbol{\hat{\theta}}_{cq,DP})-\frac{1}{n}\sum_{i\in\mathcal{I}_0}\nabla f_l(\boldsymbol{X}_i,\boldsymbol{\hat{\theta}}_{cq,DP})\}^2+ns_2^2.$$
For $\boldsymbol{h}_j^{(1)}$, since 
\begin{align}\label{va1}
	&\{\nabla^2 F_{j}(\boldsymbol{\hat{\theta}}_{cq,DP})\}^{-1}\widehat{\nabla F_{cq}^{DP}}(\boldsymbol{\hat{\theta}}_{cq,DP})\nonumber
	\\=&\{\nabla^2 F_{j}(\boldsymbol{\hat{\theta}}_{cq,DP})\}^{-1}\{\nabla^2 F_{j}(\boldsymbol{\hat{\theta}}_{cq,DP})\}\{\nabla^2 F_{j}(\boldsymbol{\hat{\theta}}_{cq,DP})\}^{-1}\widehat{\nabla F_{cq}^{DP}}(\boldsymbol{\hat{\theta}}_{cq,DP}),
\end{align}
  the variance of $\sqrt{n}h_{jl}^{(1)}$ can be estimated as:
 \begin{align}\label{va2}
 	&\frac{1}{n}\sum_{i\in\mathcal{I}_0}[\{\nabla^2 F_{0}(\boldsymbol{\hat{\theta}}_{cq,DP})\}_{l\cdot}^{-1}\{\nabla^2 f(\boldsymbol{X}_i,\boldsymbol{\hat{\theta}}_{cq,DP})\}\{\nabla^2 F_{0}(\boldsymbol{\hat{\theta}}_{cq,DP})\}^{-1}\widehat{\nabla F_{cq}^{DP}}(\boldsymbol{\hat{\theta}}_{cq,DP})]^2 \nonumber
 	\\&-\left[\frac{1}{n}\sum_{i\in\mathcal{I}_0}\{\nabla^2 F_{0}(\boldsymbol{\hat{\theta}}_{cq,DP})\}_{l\cdot}^{-1}\{\nabla^2 f(\boldsymbol{X}_i,\boldsymbol{\hat{\theta}}_{cq,DP})\}\{\nabla^2 F_{0}(\boldsymbol{\hat{\theta}}_{cq,DP})\}^{-1}\widehat{\nabla F_{cq}^{DP}}(\boldsymbol{\hat{\theta}}_{cq,DP})\right]^2 \nonumber
  \\&+ns_{3,0}^2,
 \end{align}
where $\{\nabla^2 F_{0}(\boldsymbol{\hat{\theta}}_{cq,DP})\}_{l\cdot}^{-1}$ denotes the $l$-th row of the matrix $\{\nabla^2 F_{0}(\boldsymbol{\hat{\theta}}_{cq,DP})\}^{-1}$.

\begin{theorem}\label{th32}
		Suppose Assumptions \ref{a1}-\ref{a3}, \ref{a52} and \ref{a7}-\ref{a12} hold. 
	    	When $p^2\log^2 n/n=o(1)$, ${\log^3 n}/{m}=o(1)$ and $\alpha_n=O(1/\log n)$,  it holds that 
	    \begin{align}\label{t32}
	    	\|\boldsymbol{\hat{\theta}}_{os,DP}-\boldsymbol{\theta^*}\|=O_p\left(\frac{\alpha_n\sqrt{p}}{\sqrt{n}}+\sqrt{\frac{p}{mn}}+\frac{p^3\log^2 n}{n^2}\right).
	    \end{align}
    \end{theorem}
\begin{remark}
	The difference between Theorem \ref{th31} and Theorem \ref{th32} is that the convergence rate ${p^{3/2}\log n}/{n}$ is increased to ${p^3\log^2 n}/{n^2}$. However, under the  rate assumptions $p^2\log^2 n/n=o(1)$ and $m\log n=o(n)$, this rate is still slower than $\sqrt{p/(mn)}$. In order to obtain the optimal convergence rate, we need to further iterate the estimator in subsection \ref{sec4.1.3}.
\end{remark}

	    \subsubsection{Quasi-Newton estimator}\label{sec4.1.3}

	    Below we explain how to use the quasi-Newton algorithm  for the second iteration. Firstly, the central processor transfers $\boldsymbol{\hat{\theta}}_{os,DP}$ back to the node machines, which compute the local gradient $\nabla F_j(\boldsymbol{\hat{\theta}}_{os,DP})$  and send 
	    \begin{align}\label{it5}
	    	\nabla F_j^{DP}(\boldsymbol{\hat{\theta}}_{os,DP},\boldsymbol{\hat{\theta}}_{cq,DP}):=\nabla F_j(\boldsymbol{\hat{\theta}}_{os,DP})-\nabla F_j(\boldsymbol{\hat{\theta}}_{cq,DP})+\boldsymbol{b}_j^{(4)},
	    \end{align} where $\boldsymbol{b}_j^{(4)}\sim\mathbf{N}(\boldsymbol{0},s_4^2\mathbf{I}_p)$,  back to the central processor. Secondly, the central processor computes DCQ estimators using the  values of $\nabla F_j^{DP}(\boldsymbol{\hat{\theta}}_{os,DP},\boldsymbol{\hat{\theta}}_{cq,DP})$ and $\nabla F_j^{DP}(\boldsymbol{\hat{\theta}}_{cq,DP})+\nabla F_j^{DP}(\boldsymbol{\hat{\theta}}_{os,DP},\boldsymbol{\hat{\theta}}_{cq,DP})$ , for the convenience, we denote them as $\widehat{\nabla F_{cq}^{DP}}(\boldsymbol{\hat{\theta}}_{os,DP},\boldsymbol{\hat{\theta}}_{cq,DP})$ and $\widehat{\nabla F_{cq}^{DP}}(\boldsymbol{\hat{\theta}}_{os,DP})$ respectively. Thirdly, the central processor  transmits them to the node machines, and the node machines  use quasi-Newton method to update the previously computed matrix $\{\nabla^2 F_j(\boldsymbol{\hat{\theta}}_{cq,DP})\}^{-1}$. 
	    Concretely speaking, the standard BFGS quasi-Newton method achieves the inverse of Hessian  matrix updating by computing  
	    \begin{align}\label{bfgs1}
	    	\{\widehat{\nabla^2 F_j}(\boldsymbol{\hat{\theta}}_{os,DP})\}^{-1}:=(\mathbf{V}^{(1)})^{\top}\{\nabla^2 F_j(\boldsymbol{\hat{\theta}}_{cq,DP})\}^{-1}\mathbf{V}^{(1)}+\rho^{(1)}(\boldsymbol{\hat{\theta}}_{os,DP}-\boldsymbol{\hat{\theta}}_{cq,DP})^{\otimes2},
	    \end{align} where $$\rho^{(1)}=[(\boldsymbol{\hat{\theta}}_{os,DP}-\boldsymbol{\hat{\theta}}_{cq,DP})^{\top}\{\widehat{\nabla F_{cq}^{DP}}(\boldsymbol{\hat{\theta}}_{os,DP},\boldsymbol{\hat{\theta}}_{cq,DP})\}]^{-1}$$ and $$\mathbf{V}^{(1)}=\mathbf{I}_p-\rho^{(1)}\widehat{\nabla F_{cq}^{DP}}(\boldsymbol{\hat{\theta}}_{os,DP},\boldsymbol{\hat{\theta}}_{cq,DP})(\boldsymbol{\hat{\theta}}_{os,DP}-\boldsymbol{\hat{\theta}}_{cq,DP})^{\top}.$$
	   	     Then, the node machine passes 
      \begin{align}\label{eqh2}
      \boldsymbol{h}_j^{(2)}:=\{\widehat{\nabla^2 F_j}(\boldsymbol{\hat{\theta}}_{os,DP})\}^{-1}\widehat{\nabla F_{cq}^{DP}}(\boldsymbol{\hat{\theta}}_{os,DP})+\boldsymbol{b}_j^{(5)}  \ {\rm{with}}\ \boldsymbol{b}_j^{(5)}\sim\mathbf{N}(\boldsymbol{0},s_{5,j}^2\mathbf{I}_p) 
      \end{align} 
      to the central processor, which computes the DCQ estimator $\boldsymbol{H}_2$ of $\boldsymbol{h}_j^{(2)}$ and ultimately computes the final estimator $\boldsymbol{\hat{\theta}}_{qN,DP}=\boldsymbol{\hat{\theta}}_{os,DP}-\boldsymbol{H}_2$.
	     
	      It is worth mentioning that $\rho^{(1)}$ and $\mathbf{V}^{(1)}$ are the same on different machines, and the only term in (\ref{bfgs1}) that varies with machines is $\{\nabla^2 F_j(\boldsymbol{\hat{\theta}}_{cq,DP})\}^{-1}$. Therefore, 
	      node machines only need to compute 
	      \begin{align}\label{it6}
	      	\boldsymbol{h}_j^{(3)}:=(\mathbf{V}^{(1)})^{\top}\{\nabla^2 F_j(\boldsymbol{\hat{\theta}}_{cq,DP})\}^{-1}\mathbf{V}^{(1)}\widehat{\nabla F_{cq}^{DP}}(\boldsymbol{\hat{\theta}}_{os,DP})+\boldsymbol{b}_j^{(5)},
	      \end{align} and the central processor can compute the term $\rho^{(1)}(\boldsymbol{\hat{\theta}}_{os,DP}-\boldsymbol{\hat{\theta}}_{cq,DP})^{\otimes2}\widehat{\nabla F_{cq}^{DP}}(\boldsymbol{\hat{\theta}}_{os,DP})$. These two algorithms above are essentially the same.
Write $\boldsymbol{h}_j^{(3)}$ as $(h_{j1}^{(3)},\cdots,h_{jp}^{(3)})^{\top}$. 
Similar to equation (\ref{va2}), the variance of $\sqrt{n}h_{jl}^{(3)}$ can be estimated by
\begin{align}\label{eq4.14}
	&\frac{1}{n}\sum_{i\in\mathcal{I}_0}[(\mathbf{V}^{(1)})^{\top}_{l\cdot}\{\nabla^2 F_{0}(\boldsymbol{\hat{\theta}}_{cq,DP})\}^{-1}\nabla^2 f(\boldsymbol{X}_i,\boldsymbol{\hat{\theta}}_{cq,DP})\{\nabla^2 F_{0}(\boldsymbol{\hat{\theta}}_{cq,DP})\}^{-1}\mathbf{V}^{(1)}\widehat{\nabla F_{cq}^{DP}}(\boldsymbol{\hat{\theta}}_{os,DP})]^2 \nonumber
	\\&-\left[\frac{1}{n}\sum_{i\in\mathcal{I}_0}(\mathbf{V}^{(1)})^{\top}_{l\cdot}\{\nabla^2 F_{0}(\boldsymbol{\hat{\theta}}_{cq,DP})\}^{-1}\nabla^2 f(\boldsymbol{X}_i,\boldsymbol{\hat{\theta}}_{cq,DP})\{\nabla^2 F_{0}(\boldsymbol{\hat{\theta}}_{cq,DP})\}^{-1}\mathbf{V}^{(1)}\widehat{\nabla F_{cq}^{DP}}(\boldsymbol{\hat{\theta}}_{os,DP})\right]^2\nonumber
	\\&+ns_{5,0}^2.
\end{align}

The computation procedures in Subsection \ref{sec4.1.3}  can be summarized below in  Algorithm \ref{alg1}. 
	\begin{algorithm}
	\caption{Robust distributed quasi-Newton estimator with privacy protection.}
	\label{alg1}
	\begin{algorithmic}[1]
		\REQUIRE The data set $\{X_1,X_2,\dots,X_N\}$ which is evenly distributed on $m+1$ machines $\{\mathcal{I}_0,\mathcal{I}_1,\mathcal{I}_2,\dots,\mathcal{I}_m\}$ with the local sample size $n$ and the central processor $\mathcal{I}_0$. A  positive integer $K$ used for DCQ estimator.
		\STATE  Each machine computes a local M-estimator ${\boldsymbol{\hat{\theta}}}_{j,DP}$, $j\in[m]_0$ and sends it to  $\mathcal{I}_0$.
		\STATE ({\bf Initial estimator}) $\mathcal{I}_0$  
		 computes $\boldsymbol{\hat{\theta}}_{cq,DP}$ by (\ref{eq31}).
		Then $\mathcal{I}_0$ transmits $\boldsymbol{\hat{\theta}}_{cq,DP}$ to each machine $\mathcal{I}_j, j\in[m]_0$.
		
		\STATE Each machine sends
		$\nabla F_{j}^{DP}(\boldsymbol{\hat{\theta}}_{cq,DP})$ to  $\mathcal{I}_0$.

		\STATE $\mathcal{I}_0$ computes $\widehat{\nabla F_{cq}^{DP}}(\boldsymbol{\hat{\theta}}_{cq,DP})$ 
		,
		and then sends $\widehat{\nabla F_{cq}^{DP}}(\boldsymbol{\hat{\theta}}_{cq,DP})$ to the local machines.
		\STATE Every machine sends $\boldsymbol{h}_j^{(1)}$ computed by (\ref{it1})  to $\mathcal{I}_0$.
		\STATE ({\bf One-stage estimator})  $\mathcal{I}_0$ 
		computes the DCQ estimator $\boldsymbol{H}_1$ based on $\boldsymbol{h}_j^{(1)}$. 
		Then $\mathcal{I}_0$ updates the parameter estimator by $\boldsymbol{\hat{\theta}}_{os,DP}=\boldsymbol{\hat{\theta}}_{cq,DP}-\boldsymbol{H}_1$ and sends it to the node machines.
		\STATE  The node machines send $\nabla F_j^{DP}(\boldsymbol{\hat{\theta}}_{os,DP},\boldsymbol{\hat{\theta}}_{cq,DP})$ defined by (\ref{it5}) to $\mathcal{I}_0$, and then $\mathcal{I}_0$ sends the corresponding DCQ estimator $\widehat{\nabla F_{cq}^{DP}}(\boldsymbol{\hat{\theta}}_{os,DP},\boldsymbol{\hat{\theta}}_{cq,DP})$ back to the node machines.
		\STATE ({\bf Quasi-Newton estimator}) Each machine sends $\boldsymbol{h}_j^{(3)}$ computed by (\ref{it6}) to $\mathcal{I}_0$, and $\mathcal{I}_0$ computes $\boldsymbol{H}_2$, which is the  DCQ estimator of $\boldsymbol{h}_j^{(2)}$ defined in (\ref{eqh2}), and then derive the estimator $\boldsymbol{\hat{\theta}}_{qN,DP}=\boldsymbol{\hat{\theta}}_{os,DP}-\boldsymbol{H}_2$. 
		\ENSURE The final estimator $\boldsymbol{\hat{\theta}}_{qN,DP}$.\\
	\end{algorithmic}
\end{algorithm}
    \begin{theorem}\label{th33}
    	Suppose Assumptions \ref{a1}-\ref{a5}, \ref{a52} and \ref{a7}-\ref{a12} hold. 
    	When $p^2\log^2 n/n=o(1)$, ${\log^3 n}/{m}=o(1)$ and $\alpha_n=O(1/\log n)$,
    	\begin{align}\label{t33}
    		\|\boldsymbol{\hat{\theta}}_{qN,DP}-\boldsymbol{\theta^*}\|=O_p\left(\frac{\alpha_n\sqrt{p}}{\sqrt{n}}+\sqrt{\frac{p}{mn}}\right).
    	\end{align}
    Moreover, if $\alpha_n=o_p(1/\sqrt{pm})$ and $p^9m\log^6 n/n^5=o_p(1)$, we have that for any constant vector $\boldsymbol{v}$ satisfying $\|\boldsymbol{v}\|=1$, 
    \begin{align*}
    	\frac{\sqrt{mn}\langle \boldsymbol{\hat{\theta}}_{qN,DP}-\boldsymbol{\theta^*},\boldsymbol{v}\rangle}{\boldsymbol{v}^{\top}\boldsymbol{\Sigma}_{cq}^K(\boldsymbol{\theta^*})\boldsymbol{v}}\stackrel{d}{\to}\mathbf{N}(0,1),
    \end{align*}
where $\boldsymbol{\Sigma}_{cq}^K(\boldsymbol{\theta^*})=\{\nabla^2F_{\mu}(\boldsymbol{\theta^*})\}^{-1}\mathbf
{V}_{g,vr}(\boldsymbol{\theta^*})\{\nabla^2F_{\mu}(\boldsymbol{\theta^*})\}^{-1}$, $\mathbf
{V}_{g,vr}(\boldsymbol{\theta^*})$ is a $p\times p$ matrix with the $(l_1,l_2)$-th entry ${{\sum_{k_1=1}^K\sum_{k_2=1}^K(\kappa_{k_1,k_2,K}^{l_1,l_2}-\kappa_{k_1}\kappa_{k_2})}\sigma_{gr,l_1}(\boldsymbol{\theta^*})\sigma_{gr,l_2}(\boldsymbol{\theta^*})}/{\{\sum_{k=1}^{K}\psi(\Delta_{k})\}^2}$ and $\sigma_{gr,l}^2(\boldsymbol{\theta^*})$ is the variance of $\nabla f_l(\boldsymbol{X},\boldsymbol{\theta^*})$.
    \end{theorem}
\begin{remark}
The quantity $\kappa_{k_1,k_2,K}^{l_1,l_2}$ in Theorem \ref{th33} is defined as follows:	let $(\xi_{l_1},\xi_{l_2})$ follow a mean-zero bivariate normal distribution with $\operatorname{Var}(\xi_{l_1})=\operatorname{Var}(\xi_{l_2})=1$ and $\operatorname{Cov}(\xi_{l_1},\xi_{l_2})=\frac{\sigma_{l_1,l_2}(\boldsymbol{\theta^*})}{{\sigma_{l_1}(\boldsymbol{\theta^*})\sigma_{l_2}(\boldsymbol{\theta^*}})}$. Then $\kappa_{k_1,k_2,K}^{l_1,l_2}=\mathbb{P}(\xi_{l_1}\leq\Delta_{k_1},\xi_{l_2}\leq\Delta_{k_2})$. It is easy to compute that when $K\geq 5$, each diagonal element of  $\mathbf
{V}_{g,vr}(\boldsymbol{\theta^*})$ in Theorem \ref{th33} is smaller than $1.1$ times that of $\mathbb{E}[\{\nabla f(\boldsymbol{X},\boldsymbol{\theta^*})\}^{\otimes2}]$, so $\boldsymbol{\Sigma}_{cq}^K(\boldsymbol{\theta^*})$ is very close to  $\boldsymbol{\Sigma}(\boldsymbol{\theta^*})=\{\nabla^2F_{\mu}(\boldsymbol{\theta^*})\}^{-1}\mathbf
{V}_{g,vr}(\boldsymbol{\theta^*})\{\nabla^2F_{\mu}(\boldsymbol{\theta^*})\}^{-1},$
which is the covariance matrix of the M-estimator.
\end{remark}
\subsection{Guarantee of 
differential privacy protection}\label{gn1}
 We first introduce privacy protection methods for the mean of random variables that obey sub-Gaussian (sub-exponential) distributions. In fact, we have the following two conclusions:
\begin{lemma}\label{DP1}
	Assume the data set $\mathbf{X}=\{\boldsymbol{X}_1,\cdots,\boldsymbol{X}_n\}$ is made up of i.i.d. vectors sampled from the mean-zero sub-Gaussian distribution with parameter $\nu$, that is, for any $t\in\mathbb{R}$ and $l\in[p]$, $\mathbb{E}\{\exp(tX_{l})\}\leq \exp(t^2\nu^2/2)$, where $X_l$ is the $l$-th entry of $\boldsymbol{X}$. Let $\mathcal{M}(\mathbf{X})=\frac{1}{n}\sum_{i=1}^{n}\boldsymbol{X}_i$. If we choose $\Delta=\frac{2\gamma\sqrt{p\log n}}{n}$, then the Gaussian mechinism in Lemma \ref{g1} is $(\varepsilon,\delta)$-DP with probability as least $1-2pn^{-\gamma^2/\nu^2}$.
\end{lemma}
	
\begin{lemma}\label{DP2}
	Assume the data set $\mathbf{X}=\{\boldsymbol{X}_1,\cdots,\boldsymbol{X}_n\}$ is made up of i.i.d. vectors sampled from the mean-zero  sub-exponential distribution with parameter $(\nu,\alpha)$, that is, for any $-\alpha^{-1}<t<\alpha^{-1}$, $\mathbb{E}\{\exp(tX_{l})\}\leq \exp(t^2\nu^2/2)$, where $X_l$ is the $l$-th entry of $\boldsymbol{X}$. Let $\mathcal{M}(\mathbf{X})=\frac{1}{n}\sum_{i=1}^{n}\boldsymbol{X}_i$.  
	If we choose $\Delta=\frac{2\gamma\sqrt{p}\log n}{n}$, then the Gaussian mechinism in Lemma \ref{g1} is $(\varepsilon,\delta)$-DP with probability as least $1-2p\max\{n^{-\gamma^2(\log n)/\nu^2},n^{-\gamma/\alpha}\}$.
\end{lemma}
	By selecting suitable $\gamma$, Lemmas \ref{DP1} and \ref{DP2} can guarantee that $(\varepsilon,\delta)$-DP holds with a large probability. In practical problems, we can add additional trimming procedure to the data set, so that the Gaussian mechanism can ensure $(\varepsilon,\delta)$-differential privacy strictly. 
	However, how to trim data is not the focus of this paper, and we do not further discuss specific details. According to the different distributions of random variables, the privacy budget of the exponential distribution may be  larger than that of the Gaussian distribution with $``\sqrt{\log n}"$, as shown in Lemmas \ref{DP1} and \ref{DP2}.

By controlling the eigenvalues of the Hessian matrix, Lemmas \ref{DP1} and \ref{DP2} can be extended to general M-estimators to derive the following theorem.
\begin{theorem}\label{l30}   Suppose Assumptions \ref{a1}-\ref{a3} and \ref{a52} hold.  
	Let $\boldsymbol{\hat{\theta}}=\operatorname{argmin}\limits_{\boldsymbol{\theta}\in \boldsymbol{\Theta}} \frac{1}{n}\sum_{i=1}^{n}f(\boldsymbol{X}_i,\boldsymbol{\theta})$  and
 $\boldsymbol{\hat{\theta}}_{DP}=\boldsymbol{\hat{\theta}}+\boldsymbol{b}$ with $\boldsymbol{b} \sim \mathbf{N}(\boldsymbol{0},s^2\mathbf{I}_p)$. Define $\Delta=\sqrt{2\log(1/\delta)}/\varepsilon$. Then for any positive constants $\gamma$ and $\gamma_0$,  when  $s={2.02\gamma\sqrt{p}\log n}\Delta/{(\lambda_sn)}$, $\boldsymbol{\hat{\theta}}_{DP}$ is $(\varepsilon,\delta)$-DP with the probability at least  $1-2p\max\{n^{-\gamma^2(\log n)/\nu_g^2},n^{-\gamma/\alpha_g}\}-o(n^{-\gamma_0})$.
\end{theorem}	
\begin{remark}
 The sub-exponential Assumption \ref{a52} can be enhanced to sub-Gaussian Assumption \ref{a5}, which can reduce the standard deviation $``s"$ of the noise from ${2.02\gamma\sqrt{p}\log n}\Delta/{(\lambda_sn)}$ to ${2.02\gamma\sqrt{p\log n}}\Delta/{(\lambda_sn)}$. The details can be found in Lemma 19 in Supplementary Materials.
\end{remark}

We have discussed the variance of Gaussian noise, including values from $s_1$ to $s_{5,j}$. Under the sub-exponential assumptions of the gradients and the Hessian matrices, we have the following Theorem \ref{l31}. Similar to Theorem \ref{l30}, when the gradients and the Hessian matrices obey sub-Gaussian distribution, the variance of the Gaussian noise can be reduced by a factor ``$\sqrt{\log n}$". The details can be found in Lemma 39 in Supplementary Materials.
\begin{theorem}\label{l31}   Suppose Assumptions \ref{a1}-\ref{a3}, \ref{a52} and \ref{a7} hold and $j\notin \mathcal{B}$. Let $\Delta=\sqrt{2\log(1/\delta)}/\varepsilon$.  Then for any positive constant $\gamma_0$, we have the following conclusions:
	
	(1)	When  $s_1={2.02\gamma_1\sqrt{p}\log n}\Delta/{(\lambda_sn)}$, $\boldsymbol{\hat{\theta}}_{j,DP}$ defined in (\ref{thetajdp}) satisfies $(\varepsilon,\delta)$-DP with the probability at least  $1-2p\max\{n^{-\gamma_1^2(\log n)/\nu_g^2},n^{-\gamma_1/\alpha_g}\}-o(n^{-\gamma_0})$.
	
	(2)	When $s_2={2\gamma_2\sqrt{p}\log n}\Delta/{n}$,  the gradient $\nabla F_j^{DP}(\boldsymbol{\hat{\theta}}_{cq,DP})$ defined in (\ref{eq4.5}) achieves $(\varepsilon,\delta)$-DP with the probability at least $1-2p\max\{n^{-\gamma_2^2(\log n)/\nu_g^2},n^{-\gamma_2/\alpha_g}\}-o(n^{-\gamma_0})$.
	
	(3)	 When  $s_{3,j}={2.02\gamma_3\sqrt{p}\log n}\|\{\nabla^2 F_{j}(\boldsymbol{\hat{\theta}}_{cq,DP})\}^{-1}\widehat{\nabla F_{cq}^{DP}}(\boldsymbol{\hat{\theta}}_{cq,DP})\|\Delta/{(\lambda_sn)}$, $\boldsymbol{h}_j^{(1)}$ defined in (\ref{it1}) is $(\varepsilon,\delta)$-DP with the probability at least $1-2p\max\{n^{-\gamma_3^2(\log n)/\nu_h^2},n^{-\gamma_3/\alpha_h}\}-o(n^{-\gamma_0})$.
	
	(4)	
	When  $s_4={2\gamma_4\sqrt{p}\log n}\|\boldsymbol{\hat{\theta}}_{os,DP}-\boldsymbol{\hat{\theta}}_{cq,DP}\|\Delta/{n}$, $\nabla F_j^{DP}(\boldsymbol{\hat{\theta}}_{os,DP},\boldsymbol{\hat{\theta}}_{cq,DP})$ defined in (\ref{it5}) is $(\varepsilon,\delta)$-DP with the probability at least $1-2p\max\{n^{-\gamma_4^2(\log n)/\nu_h^2},n^{-\gamma_4/\alpha_h}\}-o(n^{-\gamma_0})$.
	
	(5)	 When $s_{5,j}=2.02\gamma_5\sqrt{p}\log n\|\mathbf{V}^{(1)}\{\nabla^2 F_j(\boldsymbol{\hat{\theta}}_{cq,DP})\}^{-1}\|\|\{\nabla^2 F_j(\boldsymbol{\hat{\theta}}_{cq,DP})\}^{-1}\mathbf{V}^{(1)}\widehat{\nabla F_{cq}^{DP}}(\boldsymbol{\hat{\theta}}_{os,DP})\|\Delta/n$,  $\boldsymbol{h}_j^{(3)}$ defined in (\ref{it6}) is $(\varepsilon,\delta)$-DP with probability at least $1-2p\max\{n^{-\gamma_5^2(\log n)/\nu_h^2},n^{-\gamma_5/\alpha_h}\}-o(n^{-\gamma_0})$.

\end{theorem}
\begin{remark}
 	Although we use the same notation in Lemma \ref{l31}, $\varepsilon$ and $\delta$ from (1)  to (5) are not necessarily equal. According to \cite{dwork2006calibrating}, the composition of $k$ queries with $(\varepsilon_i,\delta_i)$-differential privacy guarantees($i=1,2,\cdots,k$), achieves at least $(\sum_{i=1}^{k}\varepsilon_i,\sum_{i=1}^{k}\delta_i)$-differential privacy. We can allocate the total privacy budget in five steps based on actual issues. When all $\varepsilon$ and $\delta$ are the same, the above Algorithm achieves $(5\varepsilon,5\delta)$-differential privacy with probability at least \begin{eqnarray*}
  p_{total}&=&1-2pm\max\{n^{-\gamma_1^2(\log n)/\nu_g^2},n^{-\gamma_1/\alpha_g}\}\\
  &&-2pm\max\{n^{-\gamma_2^2(\log n)/\nu_g^2},n^{-\gamma_2/\alpha_g}\}-2pm\max\{n^{-\gamma_3^2(\log n)/\nu_h^2},n^{-\gamma_3/\alpha_h}\}\\
  &&-2pm\max\{n^{-\gamma_4^2(\log n)/\nu_h^2},n^{-\gamma_4/\alpha_h}\}-2pm\max\{n^{-\gamma_5^2(\log n)/\nu_h^2},n^{-\gamma_5/\alpha_h}\}-o(mn^{-\gamma_0}).
  \end{eqnarray*} By adjusting the variance of the Gaussian noise in Theorem \ref{l31}, the constants  $\gamma_1$ to $\gamma_5$ can be large enough, which can guarantee the probability $p_{total}$ is close enough to $1$. Moreover, with probability at least $p_{total}$, the total privacy budget has a tighter upper bound according to Corollary \ref{la32} with $k=5$, which is mainly useful when $\varepsilon$ is small.
\end{remark}
According to Theorem 3.2 in \cite{kairouz2015composition}, we have the following corollary:
\begin{coro}\label{la32}
	Let $s_1$ to $s_{5,j}$ be defined as Theorem  \ref{l31}. For any $\varepsilon>0$, $\delta\in[0,1]$ and any slack parameter $\tilde{\delta}\in[0,1]$,
	the class of $(\varepsilon,\delta)$-differentially private mechanisms satisfy $(\tilde{\varepsilon}_{\tilde{\delta}},1-(1-\delta)^k(1-\tilde{\delta}))$-differential privacy under $k$-fold adaptive composition, where 
	\begin{align*}
	\tilde{\varepsilon}_{\tilde{\delta}}=\min\left\{k\varepsilon,\frac{(e^{\varepsilon}-1)k\varepsilon}{e^{\varepsilon}+1}+\varepsilon\sqrt{2k\log\left(e+\frac{\sqrt{k\varepsilon^2}}{\tilde{\delta}}\right)},
	\frac{(e^{\varepsilon}-1)k\varepsilon}{e^{\varepsilon}+1}+\varepsilon\sqrt{2k\log\left(\frac{1}{\tilde{\delta}}\right)}\right\}.
	\end{align*}
\end{coro}
\begin{remark}
In our research, we use the Gaussian mechanism instead of the Laplace mechanism to protect privacy. This choice is primarily due to the need to leverage the asymptotic normality of the local estimator when applying the DCQ estimator. When the perturbation term added for privacy protection follows a normal distribution, the local estimator, after the addition of the perturbation term, also adheres to a normal distribution according to the convolution formula. This approach is simpler than handling the combination of Laplace and normal distributions.
	\end{remark}

\subsection{The cases with unreliable central processor}\label{UCP}
In previous discussions, we always assumed that the central processor was not a Byzantine machine and had its own sample data. However, in practical scenarios, the central processor may only perform computational tasks without storing data, or its data may exhibit significant heterogeneity compared to that of other machines. In these situations, using the samples from the central processor to estimate the variance of local estimators would be unreliable, leading to a decrease in the effectiveness of the DCQ estimator. Since median estimation does not require variance information, we can use the median instead of the DCQ estimator for iteration.

However, for crucial statistics such as gradients, which significantly affect the final estimator, we prefer more precise estimation methods than the median. We may use node machines to compute the variance used for the DCQ estimator, but we need to ensure privacy protection when transmitting variances to the central processor. Here, we show how to calculate $\widehat{\nabla F_{cq}^{DP}}(\boldsymbol{\hat{\theta}}_{cq,DP})$ as an example. The details of privacy protection for variances used for DCQ estimator are provided in Theorem \ref{DP3}.

\begin{theorem}\label{DP3}
	Assume $X_1,\cdots,X_n$ are i.i.d. random variables sampled from the mean-zero sub-Gaussian distribution with parameter $\nu$, that is, for any $t\in\mathbb{R}$, $\mathbb{E}\{\exp(tX)\}\leq \exp(t^2\nu^2/2)$. Let $\mathcal{M}(\{X_1,\cdots,X_n\})=\frac{1}{n}\sum_{i=1}^{n}(X_i-\frac{1}{n}\sum_{i=1}^{n}X_i)^2$. If we choose $\Delta=\frac{4\gamma\log n+1}{n}$ with $\gamma\geq 1$, then the Gaussian mechinism in Lemma \ref{g1} is $(\varepsilon,\delta)$-DP with probability as least $1-8n^{-\gamma/\nu^2}$.
\end{theorem}
Replace all the DCQ estimators in Algorithm \ref{alg1} with the corresponding median estimators except for $\widehat{\nabla F_{cq}^{DP}}(\boldsymbol{\hat{\theta}}_{cq})$. To achieve privacy protection for a $p$-dimensional vector, we need to replace $(\varepsilon,\delta)$ in Lemma \ref{DP3} with $(\varepsilon/p,\delta/p)$. Before estimating the DCQ estimator $\widehat{\nabla F_{cq}^{DP}}(\boldsymbol{\hat{\theta}}_{cq})$, we need to estimate the variance of $\sqrt{n}\nabla F_1^{DP}(\boldsymbol{\hat{\theta}}_{cq,DP})$ by using the data on each node machine. Let the $j$-th node machine compute $$\hat{\sigma}_{g,jl}^2:=\frac{1}{n}\sum_{i\in\mathcal{I}_j}\{\nabla f_l(\boldsymbol{X}_i,\boldsymbol{\hat{\theta}}_{cq,DP})-\frac{1}{n}\sum_{i\in\mathcal{I}_j}\nabla f_l(\boldsymbol{X}_i,\boldsymbol{\hat{\theta}}_{cq,DP})\}^2,$$
and send $\hat{\sigma}_{g,jl}^2+b_{jl}^{(6)}$ to the central processor, where $b_{jl}^{(6)}\sim\mathbf{N}(0,s_6^2)$ with $s_6=\sqrt{2}\gamma_6p(4\log n+1)\sqrt{\log(1.25p/\delta)}/(n\varepsilon)$. The central processor  computes $\hat{\sigma}_{g,med,l}^2=\operatorname{med}\{\hat{\sigma}_{g,jl}^2+s_6^2,j\in[m]\}$, and this variance transmission achieves $(\varepsilon,\delta)$-DP with probability at least $1-8pn^{-\gamma_6/\nu^2}$.
Then $\widehat{\nabla F_{cq}^{DP}}(\boldsymbol{\hat{\theta}}_{cq})$ can be computed by $(\ref{eq21})$.

\section{Numerical Studies}

In this section, we construct some experiments on synthetic data and the MNIST real data set to examine the finite sample performance of the proposed methods. 

\subsection{Synthetic Data}
In this subsection, we perform logistic and Poisson regressions to evaluate the effectiveness and robustness of the proposed method. Throughout the simulations, we set $K$ in (\ref{eq21}) used for the DCQ estimator as $10$ and design the proportion $\alpha$ of Byzantine machines to be $\alpha=0$ and $\alpha=10\%$, representing normal and Byzantine settings, respectively. In Byzantine settings, we generate the Byzantine machines by scaling attacks based on $-3$ times normal value.  That is to say,  the statistics transmitted by the node machines to the central processor are $-3$ times the correct statistics. To demonstrate the effectiveness of the two iterations in our quasi-Newton algorithm, we compute the Mean of Root Squared Errors (MRSE) of $\boldsymbol{\hat{\theta}}_{vr,DP}$, $\boldsymbol{\hat{\theta}}_{os,DP}$, and $\boldsymbol{\hat{\theta}}_{qN,DP}$ separately over 100 simulations.

In Byzantine settings, we generate the Byzantine machines by scaling attacks based on $-3$ times normal value. That is to say, the statistics transmitted by the node machines to the central processor are $-3$ times the correct statistics.  To demonstrate the effectiveness of the two iterations in our quasi-Newton algorithm, we compute the Mean of Root Squared Errors (MRSE) of $\boldsymbol{\hat{\theta}}_{vr,DP}$, $\boldsymbol{\hat{\theta}}_{os,DP}$ and $\boldsymbol{\hat{\theta}}_{qN,DP}$ separately over 100 simulations.

{\bf Experiment 1: Logistic regression model.}
Consider  the following model:
\begin{align}\label{log2}
	Y\sim \operatorname{Bernoulli}(p_0)\ \  \text{with}\ \ p_0=\frac{\exp(\boldsymbol{X}^{\top}\boldsymbol{\theta^*})}{1+\exp(\boldsymbol{X}^{\top}\boldsymbol{\theta^*})},
\end{align}
where $Y\in\{0,1\}$ is a binary response variable, $\boldsymbol{X}\in\mathbb{R}^p$ follows the multivariate normal distribution $\mathbf{N}(\boldsymbol{0},\boldsymbol{\Sigma}_T)$ with $\boldsymbol{\Sigma}_T$ being a symmetric Toeplitz matrix with the $(i,j)$-entry $(0.6)^{|i-j|}$, $i,j\in[q]$,  $\boldsymbol{\theta^*}\in\mathbb{R}^p$ is the target parameter  and $\boldsymbol{\theta^*}=p^{-1/2}(1/2,1/2\cdots,1/2)^{\top}$.  We set the dimension of $\boldsymbol{X}$ to be $p=10$ and $p=20$ respectively.  For some simple computations and Monte Carlo estimates, we take $\gamma_1=\gamma_2=\gamma_3=\gamma_4=\gamma_5=2$.

To illustrate the impact of privacy budget on the accuracy of the parameter estimator, we consider the algorithms which achieve  
$(\varepsilon,0.05)$-DP, where $\varepsilon$ takes $4,6,8,10,12,14,16,18$, $20,30,40$ and $50$ respectively.
Since our algorithm needs to transfer $5$ vectors totally,  we add Gaussian noise to achieve $(\varepsilon/5,0.01)$-DP for each vector. 

\begin{figure}[htb!]
    \centering
    \includegraphics[width=1\linewidth]{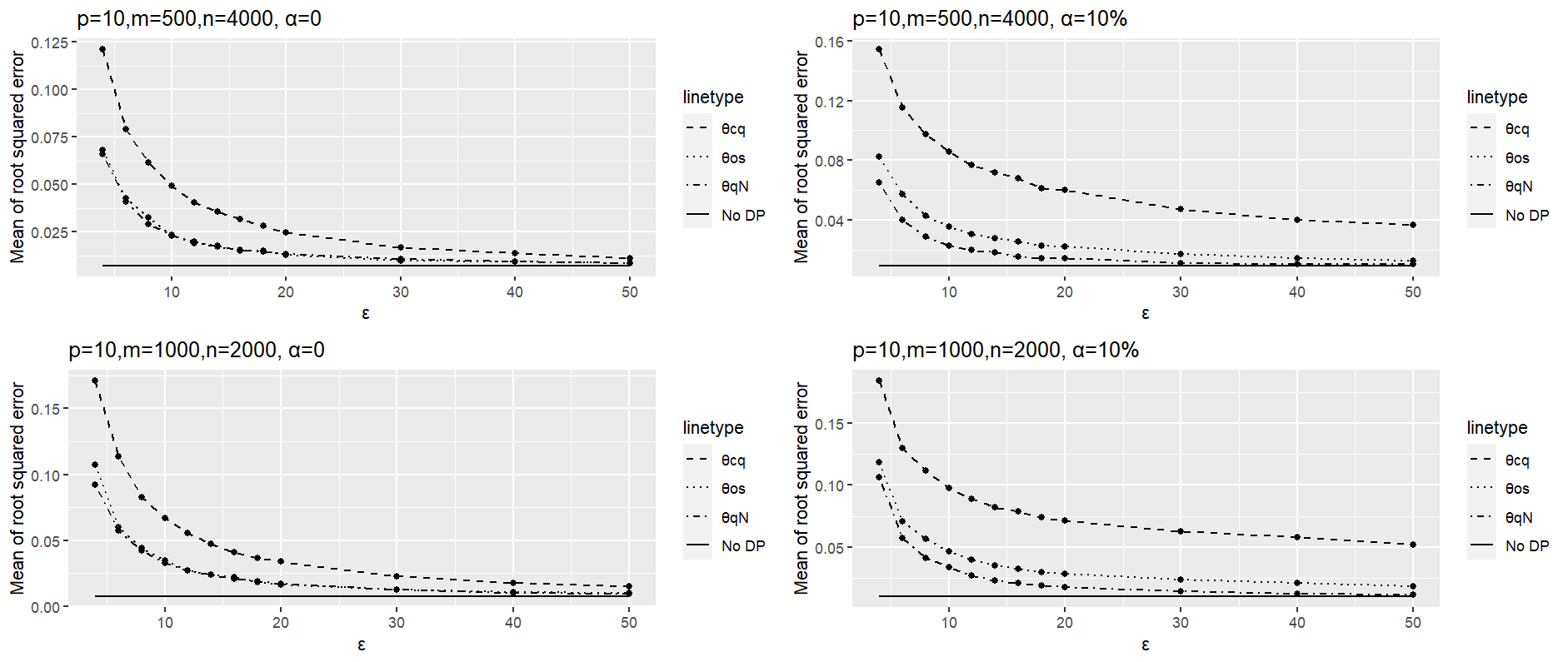}
    \caption{Logistic regression: $\varepsilon$ varies from 4 to 50, $\delta=0.05$, $p=10$, $m=500$ or $1000$, $n=4000$ or $2000$, total sample size $N=2000000$, Byzantine machine proportion $\alpha=0$ or $10\%$.}
    \label{Log0}
\end{figure}

\begin{figure}[htb!]
    \centering
    \includegraphics[width=1\linewidth]{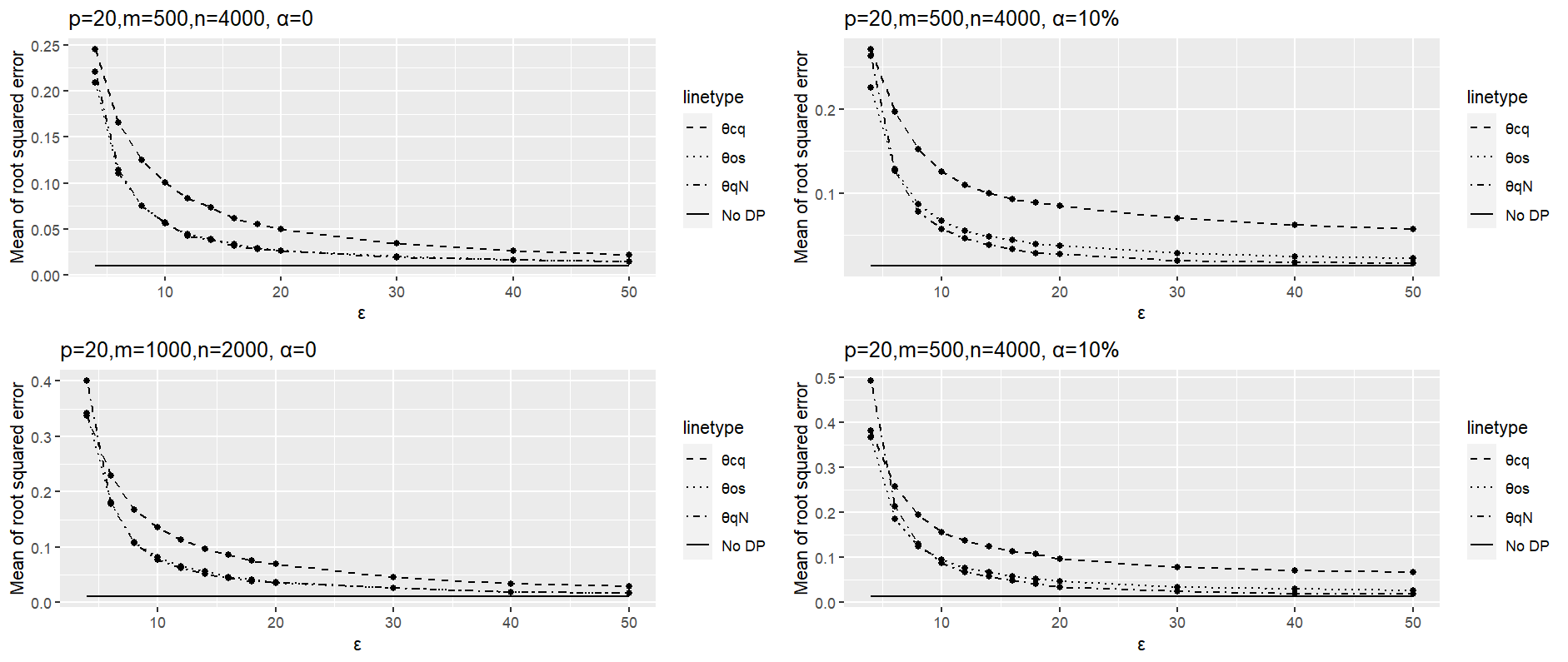}
    \caption{Logistic regression: $\varepsilon$ varies from 4 to 50, $\delta=0.05$, $p=20$, $m=500$ or $1000$, $n=4000$ or $2000$, total sample size $N=2000000$, Byzantine machine proportion $\alpha=0$ or $10\%$.}
    \label{Log1}
\end{figure}

Figures \ref{Log0} and \ref{Log1} show the line graphs to  show the variation of MRSE against $\varepsilon$, where the total sample size is $N=mn=2000000$ with the number of machines $m$ to be 500 and 1000, respectively. The dash line, dot line, and dot-dash line correspond to three estimators, $\boldsymbol{\hat{\theta}}_{vr,DP}$, $\boldsymbol{\hat{\theta}}_{os,DP}$ and $\boldsymbol{\hat{\theta}}_{qN,DP}$,  respectively. To better illustrate  the impact of noise on parameter estimation accuracy, we draw a solid line which corresponds to the quasi-Newton estimator  without privacy protection. It can be seen that the MRSE of $\boldsymbol{\hat{\theta}}_{vr,DP}$ is much larger than $\boldsymbol{\hat{\theta}}_{os,DP}$ and $\boldsymbol{\hat{\theta}}_{qN,DP}$, and the MRSE of $\boldsymbol{\hat{\theta}}_{qN,DP}$ is slightly smaller than $\boldsymbol{\hat{\theta}}_{os,DP}$. When there exists 10\% Byzantine machines, the effect of two iterations in our quasi-Newton algorithm is more pronounced. When the privacy budget $\varepsilon$ exceeds 20, the MRSE curve begins to be flat and the MRSE curve of $\boldsymbol{\hat{\theta}}_{qN,DP}$ is very close to the black line. 
Figures \ref{Log0} and \ref{Log1} show that taking $\varepsilon$ as 20 to 30 may be a good choice that balances privacy protection and estimation accuracy. 


\begin{figure}[htb!]
    \centering
    \includegraphics[width=1\linewidth]{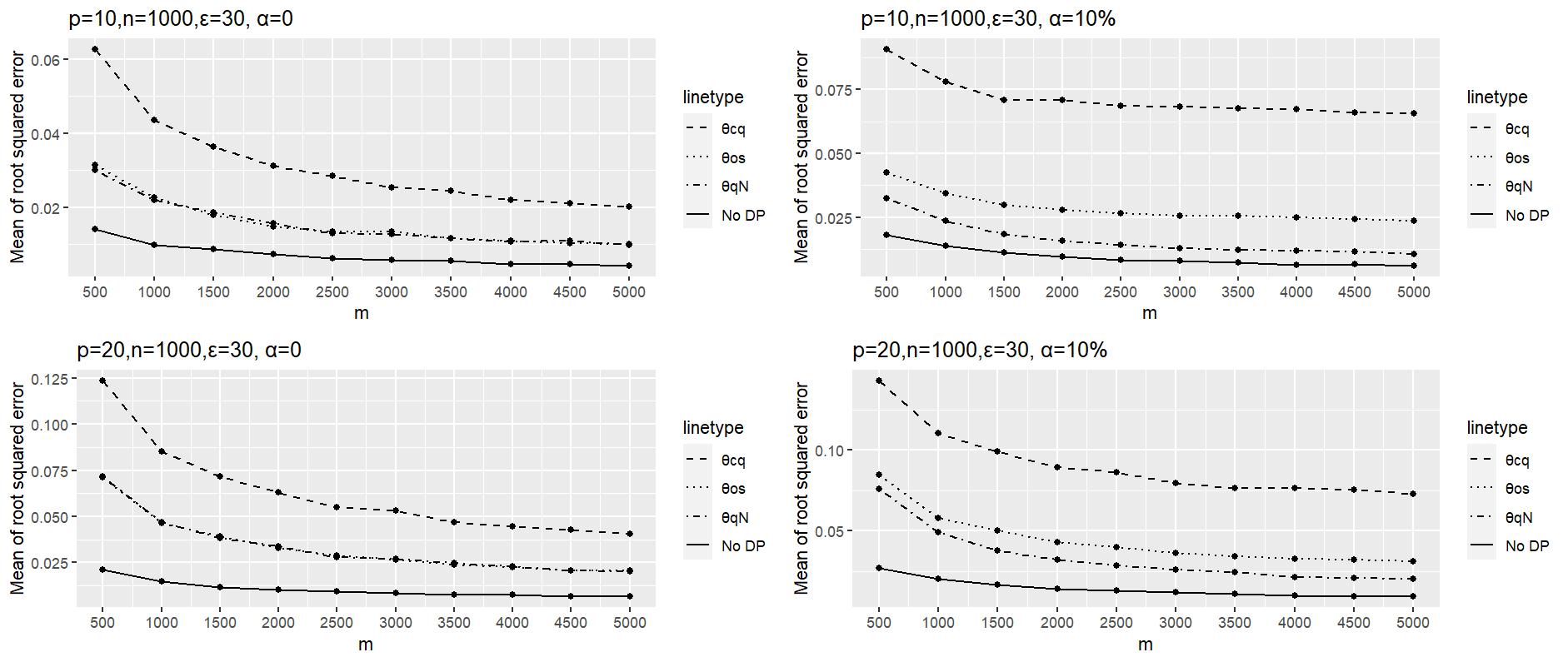}
    \caption{Logistic regression: $m$ varies from 500 to 5000, $p=10$ or $20$, $n=1000$, Byzantine machine proportion $\alpha=0$ or $10\%$, $\varepsilon=30$, $\delta=0.05$.}
    \label{Log2}
\end{figure}

Next, we consider the variation of the MRSE against the number  $m$ of machines. Consider the sample size $n=1,000$ on each node machine and  the number $m$ of machines  varies from $500$ to $5,000$. The results for $p=10$ and $20$ are shown in Figure \ref{Log2}. The MRSE gradually decreases with the increase of $m$, but slows down the decreasing rate significantly after $m$ exceeds $2000$.


{\bf Experiment 2: Poisson regression model.}
Consider the Poisson regression model:
\begin{align*}
Y\sim \operatorname{Poisson}(\boldsymbol{\lambda}) \ \text{where} \ \boldsymbol{\lambda}=\exp(\boldsymbol{X}^{\top}\boldsymbol{\theta}),
\end{align*}
where $\boldsymbol{\theta}=p^{-1/2}(1/2,1/2,\ldots,1/2)^{\top}$ is the target parameter, and $\boldsymbol{X}$ is a $p$-dimensional independent vector generated by a truncated normal 
distribution. Concretely speaking, we generate $\boldsymbol{X}_i$, for $i\in[N]$, from the normal distribution $\mathbf{N}
(\boldsymbol{0},\mathbf{\Sigma}_T)$ with $\mathbf{\Sigma}_T=(0.6)^{|i-j|}$, and regenerate $\boldsymbol{X}_i$ if $|\boldsymbol{X}_i^{\top}\boldsymbol{\theta}|>1$. In fact, more than $90\%$ of $\boldsymbol{X}_i$ satisfy $|\boldsymbol{X}_i^{\top}\boldsymbol{\theta^*}|\leq 1$, so this distribution is close to the general normal distribution.

\begin{figure}[htb!]
    \centering
    \includegraphics[width=1\linewidth]{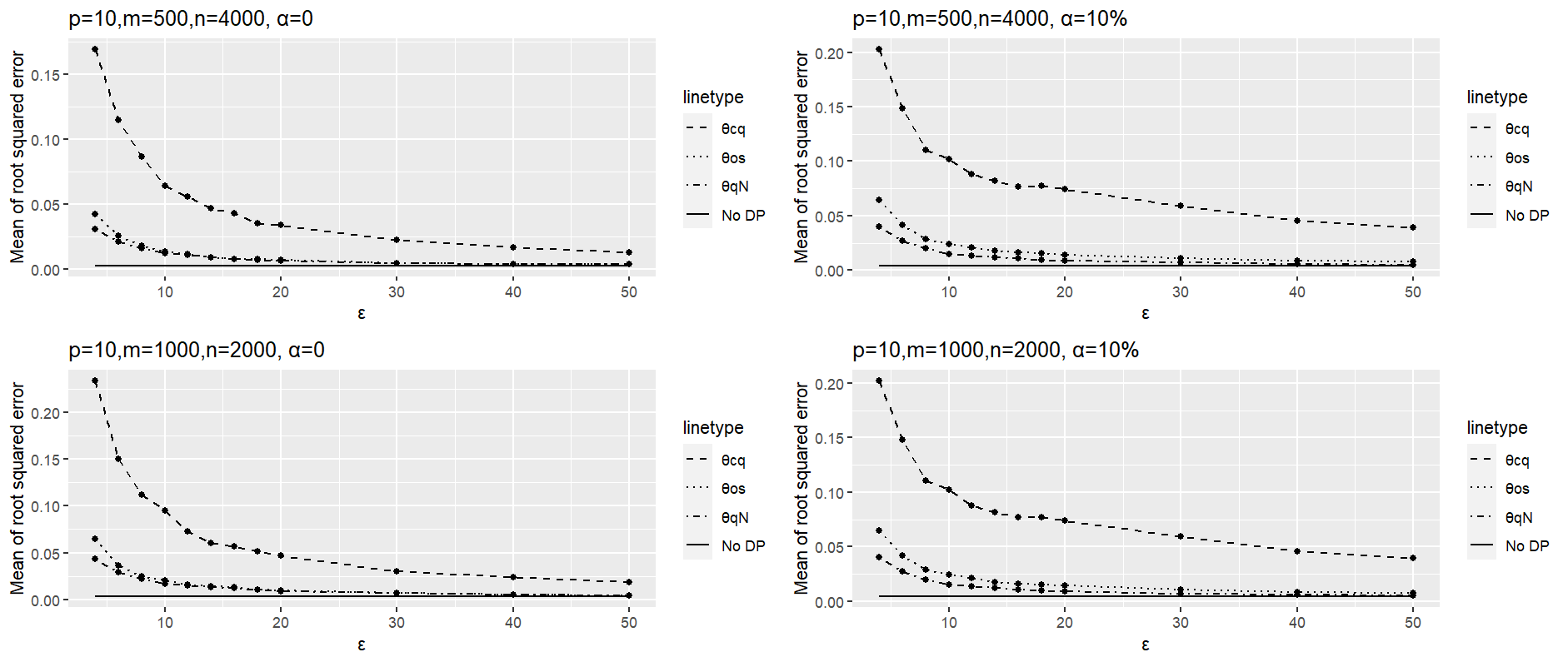}
    \caption{Poisson regression: $\varepsilon$ varies from 4 to 50, $\delta=0.05$, $p=10$, $m=500$ or $1000$, $n=4000$ or $2000$, total sample size $N=2000000$, Byzantine machine proportion $\alpha=0$ or $10\%$.}
    \label{Poi0}
\end{figure}

\begin{figure}[htb!]
    \centering
    \includegraphics[width=1\linewidth]{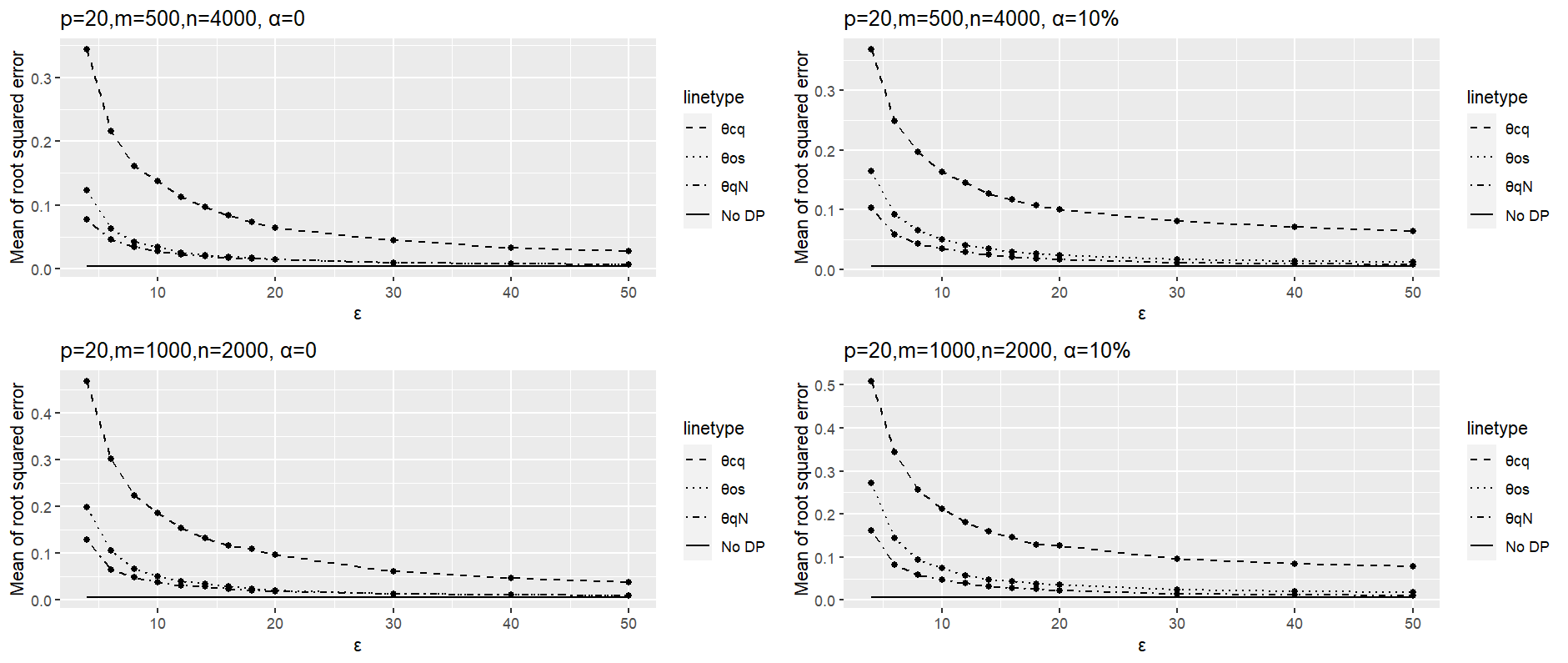}
    \caption{Poisson regression: $\varepsilon$ varies from 4 to 50, $\delta=0.05$, $p=20$, $m=500$ or $1000$, $n=4000$ or $2000$, total sample size $N=2000000$, Byzantine machine proportion $\alpha=0$ or $10\%$.}
    \label{Poi1}
\end{figure}

\begin{figure}[htb!]
    \centering
    \includegraphics[width=1\linewidth]{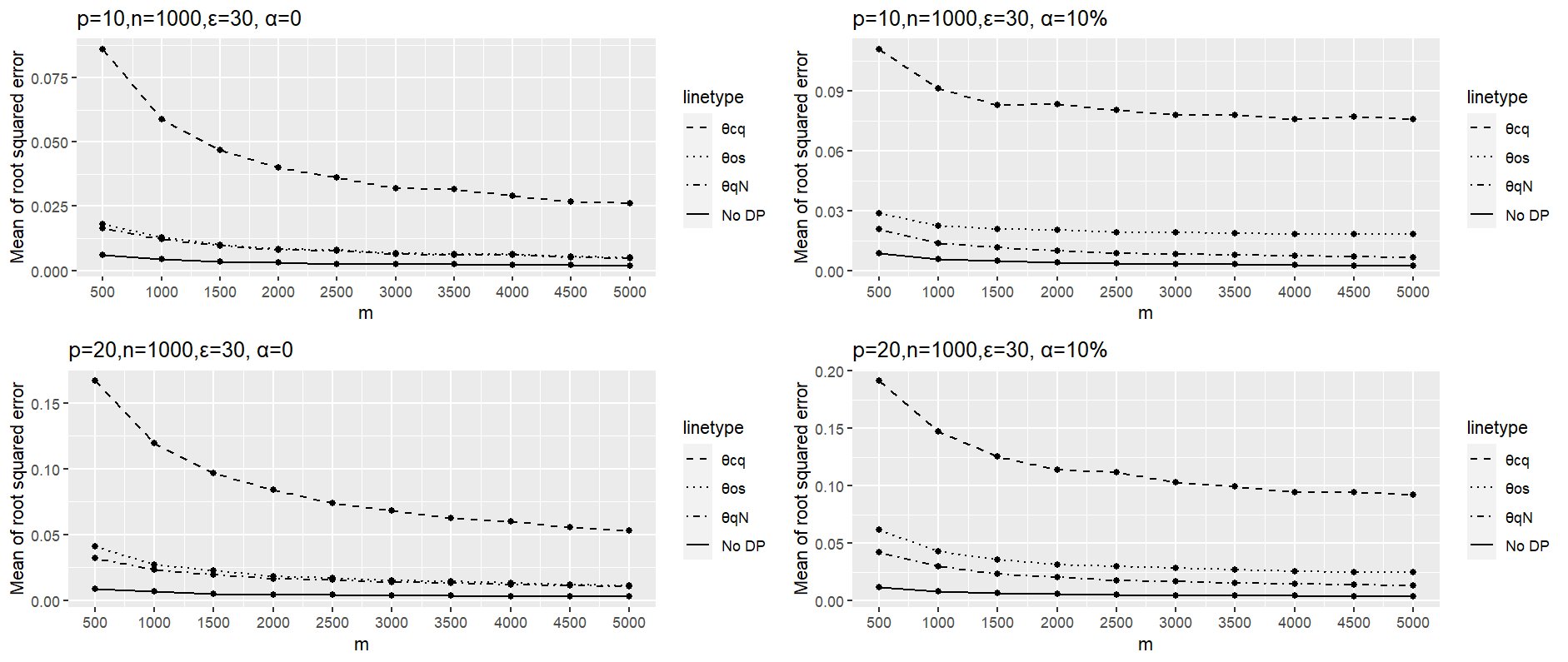}
    \caption{Poisson regression: $m$ varies from 500 to 5000, $p=10$ or $20$, $n=1000$, Byzantine machine proportion $\alpha=0$ or $10\%$, $\varepsilon=30$, $\delta=0.05$.}
    \label{Poi2}
\end{figure}

	
Figures \ref{Poi0} and \ref{Poi1} respectively show the variation of the MRSE associated with $p=10$ and $p=20$, where $\varepsilon=4,6,8,10,12,14,16,18,20,30,40,50$ and $\delta=0.05$, and Figure \ref{Poi2} displays the variation of the MRSE against the number $m$ of machines.  

The following conclusions can be drawn from Figures \ref{Poi0} to \ref{Poi2}. First, for fixed values of $(m,n)$ and $(\varepsilon,\delta)$, as the parameter dimension $p$ and the Byzantine machine proportion $\alpha$ increase, the effect of two rounds of iterations in our quasi-Newton algorithm becomes more pronounced, with the first iteration ($\boldsymbol{\hat{\theta}}_{vr,DP}$ to $\boldsymbol{\hat{\theta}}_{os,DP}$) showing a significantly greater impact than the second iteration ($\boldsymbol{\hat{\theta}}_{os,DP}$ to $\boldsymbol{\hat{\theta}}_{qN,DP}$). Additionally, when $p$ and $(m,n)$ are fixed, the MRSE decreases significantly as $\varepsilon$ increases, but when $\varepsilon$ exceeds 30, the MRSE closely approaches the quasi-Newton estimator without privacy protection (represented by the solid line in Figure \ref{Poi1}). Finally, when $p$, $n$, and $(\varepsilon,\delta)$ are fixed, the MRSE gradually decreases with an increase in $m$, although the rate of decrease slows when $m$ exceeds 2,000.

    \subsection{ A real data example}
    
        The data set is from the NIST (National Institute of Standards and Technology) database, accessible from \url{http://yann.lecun.com/exdb/mnist/}.  It contains 60,000 images and their labels, and each image is a monochrome depiction of a handwritten digit from 0 to 9, displayed against a black background with white digits, and the pixel values span from 0 to 255.

We focus on the digits 6, 8, and 9, which are recognized as difficult to differentiate. Our objective is to train three logistic classifiers to distinguish between the digit pairs 6 and 8, 6 and 9, and 8 and 9, respectively. Initially, we exclude variables where 75\% of the observations are zero. We then standardize the samples and apply the Lasso-logistic regression method described in \cite{2010Regularization} to refine the remaining variables. Finally, we select 6, 5, and 8 significant variables for the classifiers of 8 and 9, 6 and 8, and 6 and 9, respectively.
    
  We select 11,760 samples for each classifier and evenly distribute them across 10, 15, and 20 machines. For the cases with 10 and 15 machines, we assign 1 Byzantine machine, while for 20 machines, we assign 2 Byzantine machines. The Byzantine machines send values to the central processor that are 3 times the true values. First, we use the training dataset to estimate the logistic regression parameters through the quasi-Newton algorithm, and then we assess the estimation accuracy on the test dataset. Take $\gamma_1 = \gamma_2 = \gamma_3 = \gamma_4 = \gamma_5 = 0.5$. The prediction accuracy rates on the test dataset are presented in Table \ref{T01}. To illustrate the impact of privacy protection mechanisms on estimation performance, we also provide the prediction accuracy on the test set without the distributed setting and privacy protection in the ``Global" column.
     

\begin{table}[htbp]
\centering
\caption{Prediction accuracy rates of “6”, “8” and “9” in MNIST using logistic regression. In Byzantine cases, the number of Byzantine machines are respectively $1$ for $m=10$ and $15$, and $2$ for $m=20$.}
\begin{tabular}{l|ll|ll|ll|l} 
\hline
        & \multicolumn{2}{l|}{m=10, n=1176} & \multicolumn{2}{l|}{~~~~~ m=15, n=784} & \multicolumn{2}{l|}{~~~~ m=20, n=588} &                           \\ 
\hline
8 and 9 & Normal  & Byzantine               & Normal  & Byzantine                    & Normal  & Byzantine                   & Global                    \\ 
\hline
$\varepsilon=5$     & 79.04\% & 78.06\%                 & 57.86\% & 68.16\%                      & 60.88\% & 64.84\%                     &                           \\
$\varepsilon=10$    & 83.21\% & 82.84\%                 & 82.51\% & 82.84\%                      & 82.60\% & 80.33\%                     & \multirow{2}{*}{83.91\%}  \\
$\varepsilon=20$     & 83.70\% & 83.49\%                 & 83.74\% & 83.62\%                      & 83.65\% & 83.43\%                     &                           \\
$\varepsilon=30$     & 83.87\% & 83.82\%                 & 83.89\% & 83.87\%                      & 83.77\% & 83.68\%                     &                           \\ 
\hline
6 and 9 & Normal  & Byzantine               & Normal  & Byzantine                    & Normal  & Byzantine                   &                           \\ 
\hline
$\varepsilon=5$      & 88.33\% & 88.08\%                 & 87.86\% & 88.12\%                      & 87.91\% & 87.60\%                     &                           \\
$\varepsilon=10$     & 88.20\% & 88.26\%                 & 88.25\% & 88.21\%                      & 88.20\% & 88.23\%                     & \multirow{2}{*}{88.21\%}  \\
$\varepsilon=20$     & 88.30\% & 88.24\%                 & 88.17\% & 88.17\%                      & 88.19\% & 88.17\%                     &                           \\
$\varepsilon=30$     & 88.28\% & 88.26\%                 & 88.15\% & 88.15\%                      & 88.18\% & 88.19\%                     &                           \\ 
\hline
6 and 8 & Normal  & Byzantine               & Normal  & Byzantine                    & Normal  & Byzantine                   &                           \\ 
\hline
$\varepsilon=5$      & 86.49\% & 85.76\%                 & 83.84\% & 82.04\%                      & 81.31\% & 78.51\%                     &                           \\
$\varepsilon=10$     & 86.68\% & 86.84\%                 & 86.86\% & 86.57\%                      & 86.64\% & 86.76\%                     & \multirow{2}{*}{86.75\%}  \\
$\varepsilon=20$     & 86.73\% & 86.74\%                 & 86.72\% & 86.72\%                      & 86.73\% & 86.71\%                     &                           \\
$\varepsilon=30$     & 86.72\% & 86.72\%                 & 86.67\% & 86.71\%                      & 86.70\% & 86.75\%                     &                           \\
\hline
\end{tabular}
\label{T01}
\end{table}
Table \ref{T01} shows that when $\varepsilon$ exceeds 20, the prediction accuracy rate is very close to the case where  the total train dataset is used to estimate the parameters regardless of the distributed nature of obtained data and privacy protection requirement. If the total sample size remains unchanged, increasing the number of machines decreases prediction accuracy, while the influence of Byzantine machines is less significant. Additionally, the larger the number of variables used in the logistic model, the greater the required privacy budget. In our example, the regression model for digits 6 and 9, which uses 5 variables, maintains high prediction accuracy on the test set when $\varepsilon$ is equal to 5. In contrast, the regression model for digits 8 and 9, which utilizes 8 variables, exhibits a decrease in prediction accuracy as $\varepsilon$ decreases from 20 to 10, with a pronounced decline occurring when $\varepsilon$ is further reduced to 5.

	\section{Discussions}
This research explores robust estimation procedures for problems involving distributed data and Byzantine machines while ensuring data privacy. Reducing the amount of information transmission, and thereby lowering transmission costs, is a key concern. In the context of privacy protection, minimizing information transmission is even more crucial, as privacy budgets increase linearly with repeated queries. Our estimation strategy employs the quasi-Newton algorithm, which leverages Hessian matrix information to perform gradient descent, thus reducing the number of information transmission rounds while maintaining the same data transmission volume per iteration as the gradient descent method. Although this study focuses on robust estimation using DCQ estimators, the proposed robust quasi-Newton privacy-preserving algorithm is also applicable to other distributed robust estimation strategies, such as those proposed by \cite{2018Byzantine, 2019Defending}, \cite{chen2017distributed}, or \cite{su2019securing}.


Several further research directions merit investigation. First, many studies, such as \cite{Lee2017} and \cite{battery2018}, focus on distributed M-estimation with sparse structures, and the proposed robust, private distributed quasi-Newton estimation method can be applied to this problem. Additionally, exploring the combination of other privacy protection mechanisms, such as $f$-differential privacy proposed by \cite{su2022dp}, with distributed quasi-Newton methods is a worthwhile endeavor. Moreover, many references consider semi-supervised  M-estimation, such as \cite{semi2021} and \cite{song2023general}, and it would be of great interest to extend our method to the semi-supervised distributed computing problem.  Finally, the privacy-preserving DCQ estimators used for iteration leverage information from other individuals, resembling the concept of transfer learning. Thus, extending our method to transfer learning with privacy-preserving DCQ estimators, as in \cite{li2022transfer}, is another promising research direction.


	\section{Assumptions}
\begin{assumption}(Parameter space)\label{a1}
	The parameter space $\boldsymbol{\Theta} \subset \mathbb{R}^p$ is a compact convex set, and $\boldsymbol{\theta^*}$ is an interior point in $\boldsymbol{\Theta}$. The $\ell_2$-radius $D=\max_{\boldsymbol{\theta}\in \boldsymbol{\Theta}}\left\|\boldsymbol{\theta}-\boldsymbol{\theta^*}\right\|_2$ is bounded.
\end{assumption}
\begin{assumption}\label{a2}(Convexity)
	The loss function $f(\boldsymbol{x},\boldsymbol{\theta})$ is convex with respect to $\boldsymbol{\theta}\in \boldsymbol{\Theta}$ for all $\boldsymbol{x}$.
\end{assumption}
\begin{assumption}\label{a3}
	(Bounded eigenvalue for Hessian matrix)	The loss function $f(\boldsymbol{x},\boldsymbol{\theta})$ is twice differentiable, and there exists two positive constants $\Lambda_s$ and $\Lambda_l$ such that $\Lambda_s\leq\Lambda_{\min}(\nabla^2 F_{\mu}(\boldsymbol{\theta^*}))\leq\Lambda_{\max}(\nabla^2 F_{\mu}(\boldsymbol{\theta^*}))\leq\Lambda_l$.
\end{assumption}
\begin{assumption}\label{a5}(Sub-Gaussian for the gradient)
	There exists a  positive constant  $v_g$ such that for any $t\in\mathbb{R}$, $l\in[p]$ and $\boldsymbol{\theta}\in B(\boldsymbol{\theta^*},\zeta)$, 
	$$\mathbb{E}\left(\operatorname{exp}\{t|\nabla f_l(\boldsymbol{X},\boldsymbol{\theta^*})-\nabla F_{\mu l}(\boldsymbol{\theta^*})|\}\right)\leq \exp\left(\frac{1}{2}v_g^2t^2\right).$$
	
\end{assumption}
\begin{assumption}\label{a52}(Sub-exponential for the gradient)
	There exist two  positive constants $\xi_g$ and $v_g$ such that for any $0<t<\xi_g^{-1}$, $l\in[p]$ and $\boldsymbol{\theta}\in B(\boldsymbol{\theta^*},\zeta)$,
	$$\mathbb{E}\left(\operatorname{exp}\{t|\nabla f_l(\boldsymbol{X},\boldsymbol{\theta^*})-\nabla F_{\mu l}(\boldsymbol{\theta^*})|\}\right)\leq \exp\left(\frac{1}{2}v_g^2t^2\right).$$
	
\end{assumption}
\begin{assumption}\label{a6}(Bounded eigenvalue)
	There exist two positive constants $\Lambda'_s$ and $\Lambda'_l$ such that
	$$\Lambda'_s\leq\|\mathbb{E}[\{\nabla f(\boldsymbol{X},\boldsymbol{\theta^*})\}^{\otimes2}]\|\leq\Lambda'_l.$$
\end{assumption}
\begin{assumption}\label{a70}(Sub-Gaussian for   the Hessian matrix)
	There exists two positive constants $\xi_h$ and $v_h$ such that for any $t\in\mathbb{R}$ and $\boldsymbol{\theta}\in B(\boldsymbol{\theta^*},\zeta)$,
	$$\mathbb{E}\left(\operatorname{exp}[t\boldsymbol{a}^{\top}\{\nabla^2f(\boldsymbol{X},\boldsymbol{\theta^*})-\nabla^2F_{\mu}(\boldsymbol{\theta^*})\}\boldsymbol{a}]\right)\leq \exp\left(\frac{1}{2}v_h^2t^2\right),$$
	where $\boldsymbol{a}$ can be any constant vector satisfying $\|\boldsymbol{a}\|=1$.
\end{assumption}
\begin{assumption}\label{a7}(Sub-exponential for   the Hessian matrix)
	There exists two positive constants $\xi_h$ and $v_h$ such that for any $0<t<\xi_h^{-1}$  and $\boldsymbol{\theta}\in B(\boldsymbol{\theta^*},\zeta)$,
	$$\mathbb{E}\left(\operatorname{exp}[t\boldsymbol{a}^{\top}\{\nabla^2f(\boldsymbol{X},\boldsymbol{\theta^*})-\nabla^2F_{\mu}(\boldsymbol{\theta^*})\}\boldsymbol{a}]\right)\leq \exp\left(\frac{1}{2}v_h^2t^2\right),$$
	where $\boldsymbol{a}$ can be any constant vector satisfying $\|\boldsymbol{a}\|=1$.
\end{assumption}
\begin{assumption}\label{a8}
	(Sub-exponential for the inner product of vectors in the Hessian matrix)
	Let $[\nabla^2 F_{\mu}(\boldsymbol{\theta^*})]^{-1}_{l\cdot}$ be the $l$-th row of $[\nabla^2 F_{\mu}(\boldsymbol{\theta^*})]^{-1}$ and $\nabla^2 f(\boldsymbol{X},\boldsymbol{\theta})_{\cdot l}$ be the $l$-th column of $\nabla^2 f(\boldsymbol{X},\boldsymbol{\theta})$.
	There exists two positive constants $t$ and $\delta$ such that for any $l_1,l_2\in[p]$, if $\boldsymbol{\theta}\in B(\boldsymbol{\theta^*},\zeta)$, $$\mathbb{E}\left[\exp(t|\langle[\nabla^2 F_{\mu}(\boldsymbol{\theta^*})]^{-1}_{l_1\cdot},\nabla^2 f(\boldsymbol{X},\boldsymbol{\theta})_{\cdot l_2}\rangle|)\right]\leq 2.$$
\end{assumption}
\begin{assumption}\label{a9}
	(Moment ratio restriction 1)\label{berry1} For $l\in[p]$, there exists a positive constant $R_{m1}$ such that
	\begin{align*}
		\frac{\mathbb{E}\left[|\langle[\nabla^2 F_{\mu}(\boldsymbol{\theta^*})]^{-1}_{l\cdot},\nabla f(\boldsymbol{X},\boldsymbol{\theta^*})\rangle|^3\right]}{\mathbb{E}\left[\langle[\nabla^2 F_{\mu}(\boldsymbol{\theta^*})]^{-1}_{l\cdot},\nabla f(\boldsymbol{X},\boldsymbol{\theta^*})\rangle^2\right]}\leq R_{m1}.
	\end{align*}
\end{assumption}
\begin{assumption}\label{a10}(Moment ratio restriction 2)
	For any $l\in[p]$, there exist two positive constants $R_{m2}$ and $\zeta$ such that for any $\boldsymbol{\theta}\in B(\boldsymbol{\theta^*},\zeta)$,
	\begin{align*}
		\frac{\mathbb{E}[|\nabla f_l(\boldsymbol{X},\boldsymbol{\theta})-\nabla F_{\mu l}(\boldsymbol{\theta})|^3]}{\mathbb{E}[\{\nabla f_l(\boldsymbol{X},\boldsymbol{\theta})-\nabla F_{\mu l}(\boldsymbol{\theta})\}^2]}\leq R_{m2}.
	\end{align*}
\end{assumption}
\begin{assumption}\label{a102}(Moment ratio restriction 3)
	For any $l_1,l_2\in[p]$, there exist two positive constants $R_{m3}$ and $\zeta$ such that for any $\boldsymbol{\theta}\in B(\boldsymbol{\theta^*},\zeta)$
	\begin{align*}
		\frac{\mathbb{E}[|\nabla^2 f_{l_1l_2}(\boldsymbol{X},\boldsymbol{\theta})-\nabla^2 F_{\mu l_1l_2}(\boldsymbol{\theta})|^3]}{\mathbb{E}[\{\nabla^2 f_{l_1l_2}(\boldsymbol{X},\boldsymbol{\theta})-\nabla^2 F_{\mu l_1l_2}(\boldsymbol{\theta})\}^2]}\leq R_{m3}.
	\end{align*}
\end{assumption}
\begin{assumption}(Sub-exponential for  the gradient)\label{a11}
	There exist two positive constants $\xi_0$ and $v_0$ such that for any $0<t<\xi_0^{-1}$ and $l\in[p]$,
	\begin{align*}
		\mathbb{E}\left[\exp\left\{\frac{t|\nabla f_l(\boldsymbol{X},\boldsymbol{\theta})-\nabla f_l(\boldsymbol{X},\boldsymbol{\theta^*})-\nabla F_{\mu l}(\boldsymbol{\theta})+\nabla F_{\mu l}(\boldsymbol{\theta^*})|}{\|\boldsymbol{\theta}-\boldsymbol{\theta^*}\|}\right\}\right]\leq \exp\left(\frac{1}{2}v_0^2t^2\right).
	\end{align*}
\end{assumption}
\begin{assumption}\label{a12}(Smooth)
	There exist two positive constants $C_H$ and $\zeta$ such that for arbitrary $\boldsymbol{\theta}_1, \boldsymbol{\theta}_2\in B(\boldsymbol{\theta^*}, \zeta)\subset \boldsymbol{\Theta}$,
	$\left\|\nabla^2 F_{\mu}(\boldsymbol{\theta}_1)-\nabla^2 F_{\mu}(\boldsymbol{\theta}_2)\right\|\leq  C_H\left\|\boldsymbol{\theta}_1-\boldsymbol{\theta}_2\right\|$.
\end{assumption}
    \begin{assumption}\label{gf}
     For the distribution function $G(\cdot)$, let $C_g$ and $C'_g$ be two positive constants. We suppose that
     
 (1) $G^{-1}(1/2)=0$;
 
 (2) For any $x_1,x_2$, $|G(x_1)-G(x_2)|\leq C_g|x_1-x_2|$;
 
 (3) There exists a positive constant $\zeta_0<1/2$ such that for any $1/2-\zeta_0<x<1/2+\zeta_0$, $|G^{-1}(x_1)-G^{-1}(x_2)|\leq C'_g|x_1-x_2|$.
\end{assumption}
\begin{remark}
	    Assumptions \ref{a1} is common in classical statistical analysis of M-estimators (e.g., \cite{van2000asymptotic}). The convex assumption \ref{a2} can be found in \cite{fan2023communication} and \cite{chen2022first}.  Assumption \ref{a3} ensures the strong local convexity of the loss function, and similar assumptions can be found in \cite{2013Communication}, \cite{Jordan2019Communication}, and \cite{tu2021variance}. 
	
	Assumption \ref{a5} (Assumption \ref{a52})  requires each entry of the gradient to follow a sub-Gaussian (sub-exponential) distribution, which is equivalent to that in  \cite{2019Defending} and \cite{tu2021variance}. The assumptions of Gaussian and exponential distribution have an impact of $``\log n"$ on the variance of the Gaussian noise term in privacy protection. Assumption \ref{a6} requires that the covariance matrix of the gradient at the true value of the parameter has bounded eigenvalues, which is a regularity condition on studying the asymptotic properties of M-estimators (see, e.g., \cite{tu2021variance}).
	
	Assumptions \ref{a70} and \ref{a7} demand that the quadratic form of the second derivative of the loss function at the true value of the parameter also obeys a sub-Gaussian (sub-exponential) distribution.  Similar to Assumption \ref{a6}, under the sub-Gaussian assumption, the noise variance can be reduced by  $``\log n"$, and we omit further discussion. Assumption \ref{a8} requires that the inner product between the row vector about the inverse of the Hessian matrix and the second partial derivative of the loss function follows a sub-exponential distribution. Since  the non-diagonal entries of the inverse of a Hessian matrix are small in many cases, this assumption is not strong.
	
	Assumptions \ref{a9}-\ref{a102} necessitate the third-order moment of some random variables to be controlled by a constant multiple of their second-order moment, a condition frequently employed when utilizing the Berry-Esseen theorem to establish the asymptotic normality (\cite{tu2021variance}).  When a random variable exhibits pseudo-independence or adheres to an elliptical distribution (see, e.g.,  \cite{Cui2018TEST}), its third-order moment can be controlled by a constant multiple of its second-order moment. Assumptions \ref{a11} and \ref{a12} represent two smoothness assumptions, analogous to Assumptions C and D in \cite{tu2021variance}.

   Assumption \ref{gf} guarantee that $G(\cdot)$ is the distribution function of a centralized random variable and satisfying some Lipschitz conditions.   Many common limit distributions satisfy Assumption \ref{gf} after centralization, such as the normal  distribution and the chi-square distribution.
 
	Throughout all assumptions, we do not explicitly assume the dimensions $p$ of the parameter \(\boldsymbol{\theta}\) or $q$ of \(\boldsymbol{X}\), with the requirements for $p$ and $q$ contained implicitly within the functions relative to the loss function and its derivatives.
\end{remark}

	\bibliographystyle{apalike}
	\bibliography{reference(d)}

\end{document}